\documentclass[lettersize,journal]{IEEEtran}
\usepackage{amsmath,amsfonts}
\usepackage{algorithmic}
\usepackage{algorithm}
\usepackage{array}
\usepackage[caption=false,font=normalsize,labelfont=sf,textfont=sf]{subfig}
\usepackage{textcomp}
\usepackage{stfloats}
\usepackage{url}
\usepackage{verbatim}
\usepackage{graphicx}
\usepackage{cite}

\usepackage{booktabs}
\usepackage{multirow}
\usepackage{makecell}
\usepackage{graphicx}
\usepackage{tcolorbox}
\usepackage{booktabs}
\usepackage[table]{xcolor}
\usepackage{makecell}
\usepackage{array}

\usepackage{listings}
\usepackage{xcolor}
\usepackage{pdfpages}

\definecolor{codegreen}{rgb}{0,0.6,0}
\definecolor{codegray}{rgb}{0.5,0.5,0.5}
\definecolor{codepurple}{rgb}{0.58,0,0.82}
\definecolor{backcolour}{rgb}{0.95,0.95,0.92}

\usepackage{listings}
\usepackage{xcolor}

\definecolor{codebg}{rgb}{0.95,0.95,0.95}
\definecolor{framegray}{rgb}{0.8,0.8,0.8}
\definecolor{numbergray}{rgb}{0.5,0.5,0.5}
\definecolor{keywordblue}{rgb}{0.16,0.32,0.75}
\definecolor{stringpurple}{rgb}{0.58,0,0.82}
\definecolor{commentgreen}{rgb}{0,0.6,0}
\definecolor{textdark}{rgb}{0.15,0.15,0.15}

\lstdefinestyle{mystyle}{
    backgroundcolor=\color{codebg},
    frame=single,
    rulecolor=\color{framegray},
    basicstyle=\ttfamily\footnotesize\color{textdark}, 
    keywordstyle=\bfseries\color{keywordblue},
    stringstyle=\color{stringpurple},
    commentstyle=\itshape\color{commentgreen},
    numbers=left,
    numberstyle=\tiny\color{numbergray},
    numbersep=10pt,
    framexleftmargin=22pt,
    xleftmargin=22pt,
    framexrightmargin=5pt,
    framextopmargin=5pt,
    framexbottommargin=5pt,
    breaklines=true,
    keepspaces=true,
    showspaces=false,
    showstringspaces=false
}

\begin{document}

% \title{PIAC: Improvement-Aware Data and Algorithm Populations \\ Co-evolution via LLM-based Mutation}
\title{Evolving Parallel Algorithm Portfolios via \\ Potential-Aware Instance Generation with LLMs}
% \title{Potential-Aware Guided LLM Instance Generation for Evolving Parallel Algorithm Portfolios}

\author{
Shaofeng Zhang,~\IEEEmembership{Student Member,~IEEE,}
Shengcai Liu,~\IEEEmembership{Member,~IEEE,}
Zhiyuan Wang,~\IEEEmembership{Student Member,~IEEE,} and 
Ke Tang,~\IEEEmembership{Fellow,~IEEE}
\thanks{Shaofeng Zhang is with the Guangdong Provincial Key Laboratory of
Brain-Inspired Intelligent Computation, Department of Computer Science and
Engineering, Southern University of Science and Technology, Shenzhen 518055,
China, and also with the Zhongguancun Academy, Beijing 100094, China (e-mail: 12445025@mail.sustech.edu.cn).
Shengcai Liu, Zhiyuan Wang, and Ke Tang are with the Guangdong
Provincial Key Laboratory of Brain-Inspired Intelligent Computation,
Department of Computer Science and Engineering, Southern University of
Science and Technology, Shenzhen 518055, China (e-mail: liusc3@sustech.edu.cn; wangzy2020@mail.sustech.edu.cn; tangk3@sustech.edu.cn).}}




\maketitle

\begin{abstract}
The Automatic Construction of Portfolios via Large Language Models (LLM-ACP) suffers from poor generalization in practical few-shot scenarios when solving complex combinatorial optimization problems. Instance and algorithm co-evolution frameworks address this by expanding the training dataset with generated hard instances on which the current algorithm portfolio underperforms, thereby enhancing generalization. However, this paradigm faces two critical limitations: evaluating instance hardness relies on high-quality reference solutions, and single-mode generation patterns limit instance diversity. To overcome these limitations, we introduce the Potential-aware Instance and Algorithm Co-evolution (PIAC) framework. Our core contribution is twofold. First, we propose \emph{potential gain}, a novel metric that eliminates the need for reference solutions. This metric estimates generalization gain by perturbing the generated algorithms and assessing their improvement potential on generated problem instances. Second, PIAC leverages LLMs to synthesize diverse instance mutators, exploring a broader region of the problem-instance space and thereby enhancing the portfolio's generalization capabilities. Given that perturbation spaces vary across different algorithms, we instantiate our framework on Greedy Constructive, Ant Colony Optimization, and Guided Local Search algorithmic backbones. Comprehensive evaluations on the Traveling Salesman Problem (TSP) and Capacitated Vehicle Routing Problem (CVRP) across six distinct data distributions demonstrate that PIAC consistently outperforms state-of-the-art LLM-ACP baselines, notably achieving a 19.76\% relative improvement for TSP Greedy Constructive portfolios.
\end{abstract}

\begin{IEEEkeywords}
Parallel algorithm portfolios, large language model, automatic algorithm design, co-evolutionary algorithm.
\end{IEEEkeywords}

\section{Introduction}
\label{sec:intro}

Parallel Algorithm Portfolios (PAPs) have emerged as a powerful paradigm for solving complex engineering optimization problems~\cite{Bernardo1997, Carla2001ap}. Inspired by the No Free Lunch theorem~\cite{Wolpert1997NoFreeLunch}, which posits that different algorithms exhibit varying performance across different problem characteristics, PAPs aim to construct a complementary set of member algorithms to enhance overall performance across the entire problem space. During inference, a PAP typically runs multiple member algorithms in parallel and returns the best solution found as the final output. In doing so, PAPs can effectively exploit the complementarity among their members while fully leveraging modern parallel computing resources, such as multi-core CPUs, to achieve superior overall performance.

Since manually constructing a high-quality PAP is non-trivial, Automated Construction of Portfolios (ACP) has been extensively studied~\cite{xu2010hydra, lindauer2016automatic}. To reduce manual effort, mainstream ACP frameworks adopt the data-driven paradigm. Given an algorithm search space (traditionally the parameter configuration space of base solvers) and a set of training instances, the standard framework iteratively explores this space to evolve candidate algorithms. Guided by their performance on the training instances, it ultimately returns a portfolio of complementary member algorithms that maximizes overall performance. Recent advances in the programming capabilities of Large Language Models (LLMs) have introduced Automatic Construction of Portfolios via LLMs (LLM-ACP) as a powerful new subfield of ACP~\cite{liu2026survey, zhang2026survey, liu2025eohs}. The fundamental distinction between LLM-ACP and ACP lies in the algorithm search space. LLM-ACP shifts the paradigm from tuning fixed parameter configurations to exploring an open-ended programmatic space, such as heuristic code snippets.

Despite this progress, ACP (including LLM-ACP) frequently encounters the \emph{few-shot} generalization challenge. In practice, only a limited number of training instances are available, and they often fail to capture the complex instance distributions encountered in real-world data~\cite{Reinelt1991tsplib, Uchoa2017cvrplib}. Consequently, portfolios constructed on such limited training sets are highly susceptible to overfitting and often fail to generalize to unseen instances. To overcome this limitation, prior studies (e.g., CEPS~\cite{tang2021ceps} and DACE~\cite{wang2025dace}) have leveraged instance-algorithm co-evolution to dynamically expand the training dataset. In this paper, we define the \textit{quality} of a problem instance as the improvement in a PAP's generalization performance achieved by incorporating it into the training dataset. To obtain such high-quality instances, existing methods actively generate adversarial or ``hard'' instances that the current portfolio struggles to solve. Specifically, these hard instances expose regions of the problem space where current algorithms underperform. Augmenting the training set with these instances drives the portfolio to improve its performance in these specific regions, thereby enhancing overall generalization.

However, adapting traditional co-evolutionary frameworks, such as CEPS and DACE, to the LLM-ACP paradigm exposes two critical limitations. First, regarding \textbf{reliance on high-quality solutions}, existing methods typically depend on near-optimal reference solutions to evaluate the quality of newly generated instances~\cite{Hemert2006, wang2024asp, Branke2011}. Specifically, these approaches evaluate an instance's quality based on its difficulty, termed the \textit{hardness metric}, and quantify this metric by calculating the performance gap between the current PAP and the reference solutions. In many practical or newly formulated scenarios, acquiring such high-quality solutions is a challenging task. Second, regarding \textbf{single-mode instance generation patterns}, prior instance generation strategies often rely on fixed, predefined operators, such as simple geometric transformations or random perturbations~\cite{tang2021ceps, wang2025dace}. This single-mode generation paradigm lacks diversity, yielding highly homogeneous instances that restrict the generalization capability of the resulting algorithm portfolio.

To overcome these limitations, we introduce the Potential-aware Instance and Algorithm Co-evolution (\textbf{PIAC}) framework. While building upon the co-evolution framework, PIAC advances the paradigm through two core innovations tailored to LLM-ACP. 
First, we propose \emph{potential gain}, a novel potential-aware metric that avoids the need for high-quality reference solutions. Rather than relying on the hardness metric to determine an instance's quality, PIAC shifts the evaluation focus to the algorithm portfolio's ``potential for improvement''. The potential gain metric operates by perturbing the designed algorithms and quantifying the performance improvement achieved by the perturbed algorithms relative to the original algorithms on newly generated instances. By directly measuring this empirical gain, \emph{potential gain} provides a surrogate signal for the portfolio's improvement potential on the given instance. Second, to enhance the diversity of generated instances, PIAC leverages the code-generation capabilities of LLMs to synthesize and evolve a population of programmatic instance mutators. By transitioning from static, predefined operators to diverse mutators, PIAC effectively expands the searchable instance space and further enhances generalization. Together, potential-aware evaluation and programmatic instance synthesis augment the training set with valuable instances, guiding portfolio evolution toward greater complementarity and better generalization. To evaluate the generalization of the algorithm portfolios, we construct the PAPs in the few-shot setting and measure their performance across six distinct data distributions for both the Traveling Salesman Problem (TSP) and the Capacitated Vehicle Routing Problem (CVRP). The main contributions of this work are summarized as follows:

\begin{itemize}
    \item \textbf{Potential-Aware Instance Evaluation.} We propose a novel \textit{potential-aware} metric to evaluate generated instances without reference solutions. By perturbing the generated algorithms, the metric estimates the performance gains achievable on each instance, thereby identifying high-quality training instances.
    
    \item \textbf{LLM-Driven Instance Evolution Mechanisms.} We leverage the code-generation capabilities of LLMs to dynamically synthesize a diverse set of instance mutators, enabling the instance generation process to explore a broader and more diverse instance space.
    
    \item \textbf{Instantiation on Multiple Algorithmic Backbones.} We apply our framework to Greedy Constructive, Ant Colony Optimization, and Guided Local Search algorithmic backbones. Comprehensive evaluations demonstrate that PIAC consistently outperforms state-of-the-art LLM-ACP baselines, notably achieving a 19.76\% relative improvement for TSP Greedy Constructive portfolios.
\end{itemize}

The remainder of this paper is organized as follows. Section~\ref{sec:llm_pap} formalizes the problem of ACP and reviews relevant prior work. Section~\ref{sec:framework} introduces the proposed PIAC framework. Section~\ref{sec:instantiations} provides algorithm-specific instantiations of the framework. Section~\ref{sec:experiments} presents a comprehensive empirical evaluation of our approach. Finally, Section~\ref{sec:conclusion} concludes the article and outlines directions for future research. Our code is available at
\textcolor{blue}{\url{https://anonymous.4open.science/r/piac-BAF3}}.

\section{Problem Definition and Related Work}
\label{sec:llm_pap}

\subsection{Problem Formulation}
Assume that the construction of a PAP is defined over the entire problem class.
Let \(x \in \Omega\) denote a problem instance, where \(\Omega\) represents the complete instance space.
The backbone $B$ specifies a complete solution procedure, such as greedy construction or ant colony optimization. Let $\mathcal{H}_B$ denote the corresponding space of heuristic functions that can be constructed by the LLM. Given an instance $x\in\Omega$ and a solver state $s$ encountered during solution construction or search, each $H\in\mathcal{H}_B$ produces a heuristic matrix $M=H(x,s)$ that guides the subsequent decision made by the backbone. Instantiating the fixed backbone $B$ with $H$ yields a complete executable algorithm, denoted by $A=B[H]$. Accordingly, the complete algorithm space under this backbone is
\begin{equation}
    \mathcal{A}
    =
    \{B[H]\mid H\in\mathcal{H}_B\}.
\end{equation}
PIAC employs the LLM to construct and evolve the heuristic component $H$ while keeping $B$ unchanged. Throughout this paper, ``algorithm'' refers to the complete executable algorithm $A=B[H]$, rather than the isolated heuristic function $H$.
A PAP is denoted by $\mathbb{A} = \{A_1, \ldots, A_K\}$, where $A_j=B[H_j]\in\mathcal{A}$ and $\mathbb{A} \subseteq \mathcal{A}$.
Given a problem instance \(x\), let \(F(A,x)\) denote the performance of algorithm \(A\) on \(x\).
Without loss of generality, we assume that smaller values indicate better performance.
Accordingly, the performance of the PAP \(\mathbb{A}\) on instance \(x\) is defined as
\begin{equation}
F(\mathbb{A},x)
=
\min_{A \in \mathbb{A}} F(A,x),
\end{equation}
which indicates that the portfolio performance on a given instance is determined by the best-performing algorithm within the PAP.

Ideally, the objective of automated PAP construction is to identify an optimal PAP, denoted by \(\mathbb{A}^\ast\), that achieves the best generalization performance over the entire problem domain:
\begin{equation}
\mathbb{A}^\ast
=
\arg\min_{\mathbb{A} \subseteq \mathcal{A},\, |\mathbb{A}|=K}
\mathbb{E}_{x \sim p(x)}
\left[
F(\mathbb{A},x)
\right],
\end{equation}
where \(p(x)\) denotes the underlying distribution over the problem instance space.

However, in practical scenarios, the true distribution over the entire instance space is usually unavailable.
Therefore, existing PAP construction methods commonly optimize the empirical performance of the PAP on a given training dataset \(\mathcal{D} \subset \Omega\):
\begin{equation}
\widehat{\mathbb A}^\ast
=
\arg\min_{\mathbb{A} \subseteq \mathcal{A},\, |\mathbb{A}|=K}
F(\mathbb{A},\mathcal{D}),
\end{equation}
where
$F(\mathbb{A},\mathcal{D})=\frac{1}{|\mathcal{D}|}\sum_{x \in \mathcal{D}}F(\mathbb{A},x)$.

This empirical objective implicitly assumes that the training dataset \(\mathcal{D}\) is sufficiently representative of the complete problem instance space \(\Omega\).
Nevertheless, in few-shot scenarios, the available training instances are often scarce and insufficient to capture the diversity and critical structural characteristics of the underlying problem distribution.
As a result, the constructed PAP may suffer from limited generalization ability when applied to unseen problem instances.

\begin{figure*}[!t]
    \centering
    \includegraphics[width=1.\textwidth]{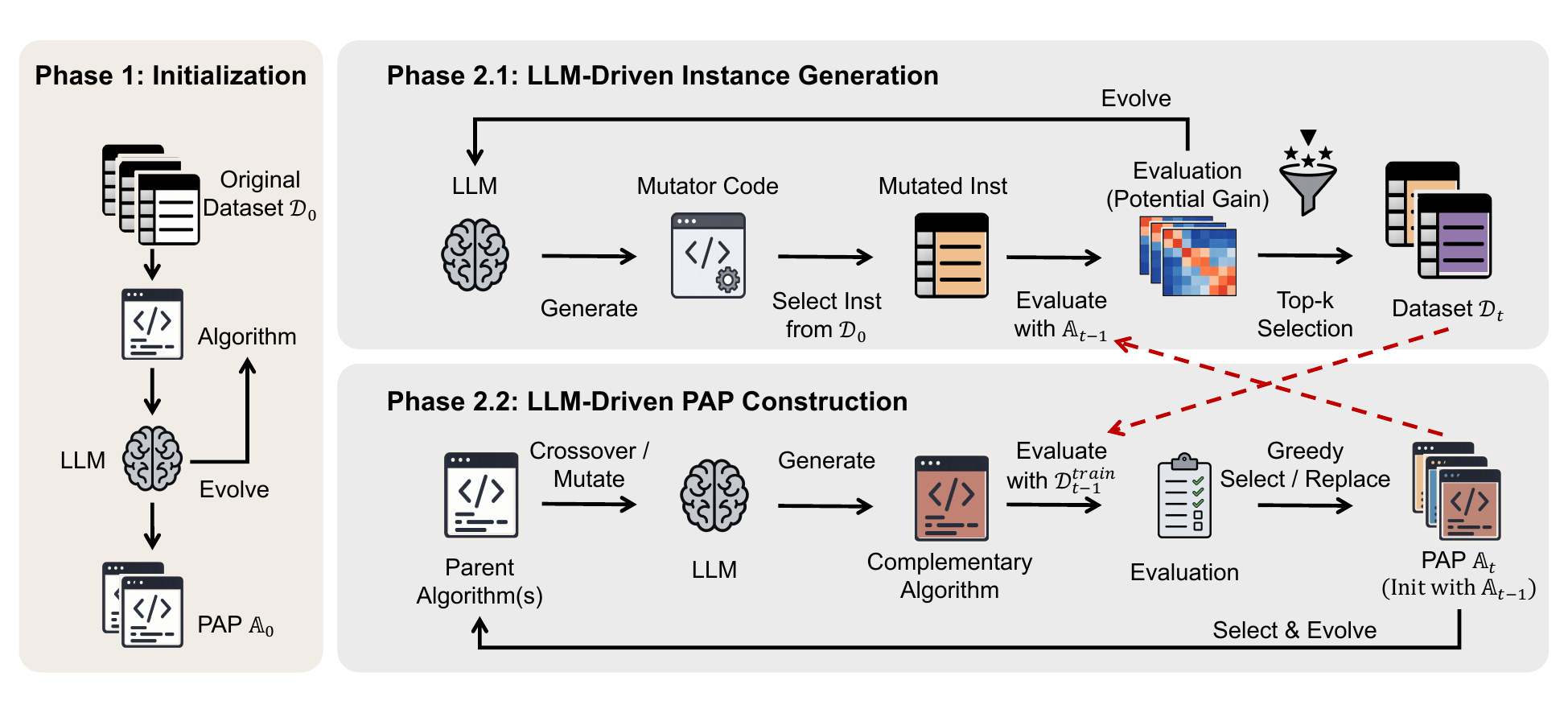}
    \vspace{-10pt}
    \caption{\textbf{PIAC Framework:} A co-evolutionary framework for LLM-driven portfolio construction. By alternating between instance synthesis via LLM-generated mutators and algorithm portfolio optimization, PIAC jointly enhances training data coverage and the generalization performance of the algorithm portfolios.}
    \label{fig:workflow}
\end{figure*}

\subsection{Existing ACP Framework}
The automated construction of algorithm portfolios (ACP) has driven significant advancements in solving hard combinatorial problems~\cite{Carla2001ap, Pang2026Balancing, Guo2026Multitree}. Existing approaches generally follow two paradigms. The first constructs a globally complementary set of member algorithms evaluated across the entire instance space, as exemplified by Hydra~\cite{xu2010hydra}, GLOBAL~\cite{lindauer2016automatic}, and PARHYDRA~\cite{lindauer2016automatic}. In contrast, the second paradigm addresses instance heterogeneity by partitioning the problem space and constructing a tailored component algorithm for each specific cluster, as demonstrated by CLUSTERING~\cite{kadioglu2010isac} and PCIT~\cite{liu2019automatic}. By yielding diverse member algorithms, these paradigms inherently facilitate automated algorithm selection (AS)~\cite{wu2024, wu2025towards, Pei2025ope}. However, traditional frameworks typically cast ACP strictly as a parameter configuration problem. Consequently, they confine the search to the predefined parameter spaces of existing solvers, fundamentally precluding the discovery of structurally novel algorithms.

To break these structural limits, recent studies leverage LLMs to automatically design heuristics or executable code for discrete optimization~\cite{wu2025survey, JiangSQLZZY25, liu2024llmeo, Xu2026EvoSpeak}. These methods generally establish a novel paradigm integrating LLMs with Evolutionary Computation (EC). In this framework, LLMs act as intelligent variation operators within an evolutionary loop to iteratively generate and refine heuristic code, as demonstrated by methods such as FunSearch~\cite{romera2023funsearch}, EoH~\cite{liu2024eoh}, ReEvo~\cite{ye2024reevo}, HSEvo~\cite{dat2024hsevo}, and MCTS-AHD~\cite{zheng2025mcts}. This paradigm, formally recognized as LLM-based Automatic Heuristic Design (LLM-AHD), has been widely applied across diverse domains, including recommender systems~\cite{liu2025autorec} and scientific discovery~\cite{Chen2025}. Building upon this foundation, the research frontier has shifted toward LLM-ACP to explicitly address instance heterogeneity. For example, InstSpecHH~\cite{zhang2024instspechh} partitions the problem space to evolve instance-specific heuristics for feature-based subclasses. Concurrently, EoH-S~\cite{liu2025eohs} directly targets portfolio synergy by employing a complementary-aware memetic search to evolve a highly cooperative heuristic set that collectively covers diverse instance distributions.

While existing LLM-ACP methods demonstrate strong performance, they predominantly rely on the assumption of abundant training data, inherently struggling with the few-shot generalization challenge in data-scarce scenarios. In traditional ACP, data scarcity is typically mitigated via co-evolutionary frameworks, such as CEPS~\cite{tang2021ceps}, GAST~\cite{liu2020gast}, and DACE~\cite{wang2025dace}. These frameworks fundamentally operate through a competitive, two-step iterative mechanism. The process begins by evolving a PAP over a given training dataset to maximize overall portfolio performance. Subsequently, the training set is augmented by generating and incorporating novel instances explicitly designed to minimize the performance of the current PAP. These two adversarial phases alternate continuously, ultimately yielding a robust parallel algorithm portfolio with strong generalization capabilities. Extending this adversarial paradigm to deep learning, ASP~\cite{wang2024asp} formulates distributional exploration as a two-player zero-sum meta-game to train best-response generators alongside neural solvers.

Despite the potential of integrating this co-evolutionary paradigm with LLM-ACP, current frameworks present two primary limitations. First, regarding the reliance on high-quality solutions, existing methods typically depend on near-optimal reference solutions to evaluate the difficulty of newly generated instances~\cite{Hemert2006, wang2024asp, Branke2011}. This dependency severely restricts the framework's applicability, making it challenging to extend to practical or newly formulated problem domains. Second, regarding single-mode instance generation, current instance generation strategies often rely on static, predefined generation operators~\cite{tang2021ceps, wang2025dace}. This unimodal generation paradigm fundamentally lacks diversity, yielding highly homogeneous instances that hinder further improvements to the generalization capability of the evolved PAP.

\section{The Proposed PIAC Framework}
\label{sec:framework}
We extend the adversarial co-evolution framework~\cite{tang2021ceps} to the LLM-ACP paradigm. While our Potential-aware Instance and Algorithm Co-evolution (PIAC) framework adopts the alternating optimization phases of CEPS to iteratively evolve both the algorithm portfolio and the training instances, its core novelty lies within the instance evolution phase. Existing co-evolution frameworks typically equate instance quality with instance hardness, inherently relying on high-quality reference solutions, and are often restricted to single-mode generation patterns. PIAC addresses these limitations by introducing the potential gain metric, which circumvents the need for reference solutions, coupled with an evolutionary search over mutator programs to expand the searchable instance space.

As illustrated in Algorithm~\ref{alg:piac}, the framework consists of two alternating phases:
\begin{enumerate}
    \item \textbf{LLM-Driven Instance Generation:} The LLM generates executable mutator code to synthesize novel instances. 
    The primary objective of this process is to maximize the overall quality of the generated instances $\max_{\mathcal{D}_t} \sum_{x \in \mathcal{D}_t} V(x; \mathbb{A}_{t-1})$, where $V(x;\mathbb{A}_{t-1})$ denotes the potential gain of generated instance $x$ with respect to the current portfolio $\mathbb{A}_{t-1}$.
    \item \textbf{LLM-Driven Algorithm Portfolio Construction:} Subsequently, the LLM evolves the heuristic portfolio to address these challenging new instances. The goal is to identify a portfolio $\mathbb{A}_t$ of size $K$ that minimizes the evaluation metric $F$ over the accumulated training set $\mathcal{D}^{train}$: $\min_{\mathbb{A}_t : |\mathbb{A}_t|=K} F(\mathbb{A}_t, \mathcal{D}^{train})$.
\end{enumerate}

To systematically present these contributions, Section~\ref{subsec:llm_data_gen} details the LLM-driven instance generation phase, encompassing both the mutator search and the potential-aware evaluation. Subsequently, Section~\ref{subsec:pap_construct} outlines the LLM-based algorithm portfolio construction process.

\begin{algorithm}[t]
\caption{PIAC}
\label{alg:piac}
\begin{algorithmic}[1]
\REQUIRE Initial dataset $\mathcal{D}_0$, algorithmic backbone $B$, portfolio size $K$, max iterations $T$, max mutator evaluations $FE_m$, max algorithm evaluations $FE_a$, number of augmented instances $N_{\mathrm{aug}}$, number of executions per mutator $N_{\mathrm{exec}}$

\ENSURE Final complementary algorithm portfolio $\mathbb{A}_T$

\STATE \textit{/* Phase 1: Initialization */}
\STATE Initialize accumulated training dataset: $\mathcal{D}^{train} \leftarrow \mathcal{D}_0$.

\STATE Construct initial portfolio $\mathbb{A}_0$ of size $K$ on $\mathcal{D}_0$ via LLM generation and greedy selection.

% \STATE Construct initial portfolio $\mathbb{A}_0$ of size $K$ by embedding LLM-generated heuristic components into $B$ and applying greedy selection.

\STATE \textit{/* Phase 2: Co-Evolution Loop */}
\FOR{$t = 1, 2, \ldots, T-1$}
    \STATE \textit{/* Step 2.1: LLM-Driven Data Generation*/}
    \STATE Initialize candidate instance set $\mathcal{X}_{cand} \leftarrow \emptyset$.
    \STATE Prompt LLM to generate initial mutator set $\mathcal{M}_{prev}$.
    \STATE \(e_m\leftarrow0, e_a\leftarrow0.\)
    
    \WHILE{$e_m < FE_m$}
        \STATE Select parent mutators from $\mathcal{M}_{prev}$.
        \label{algline:mutator_parent_selection}
        
        \STATE Prompt LLM to generate child mutators $\mathcal{M}_{child}$.
        \label{algline:child_mutator_generation}
        
        \FOR{\textbf{each} child mutator $m \in \mathcal{M}_{child}$}
            \STATE Execute $m$ for $N_{\mathrm{exec}}$ times, independently sampling base instances $x_{base} \sim \mathcal{D}_0$ each time to synthesize $\mathcal{X}_{new}$.
            \label{algline:mutate_base_instances}
            
            \STATE Evaluate the potential gain $V(x; \mathbb{A}_{t-1})$ for instances in $\mathcal{X}_{new}$ via Eq.~\eqref{eq:perturb_gain}.
            \STATE Update fitness of $m$ via Eq.~\eqref{eq:mutator_fitness}.
            \STATE Add the evaluated instances $\mathcal{X}_{new}$ to $\mathcal{X}_{cand}$.
        \ENDFOR
        \STATE $\mathcal{M}_{prev} \leftarrow \mathcal{M}_{child}$, $e_m \leftarrow e_m + |\mathcal{M}_{child}|$.
    \ENDWHILE
    \STATE Select top-$N_{\mathrm{aug}}$ instances from $\mathcal{X}_{cand}$ with maximum $V(x; \mathbb{A}_{t-1})$ to form $\mathcal{D}_t$.
    \label{algline:select_informative_instances}
    
    \STATE $\mathcal{D}^{train} \leftarrow \mathcal{D}^{train} \cup \mathcal{D}_t$.
    \label{algline:update_training_dataset}
    
    \STATE \textit{/* Step 2.2: LLM-Driven Algorithm Construction */}
    \STATE Initialize current portfolio $\mathbb{A}_t \leftarrow \mathbb{A}_{t-1}$.
    \label{algline:init_current_portfolio}

    \WHILE{$e_a < FE_a$}
        \label{algline:algorithm_evolution_loop}
        \STATE Select parent algorithms from $\mathbb{A}_t$ for either \textbf{Crossover} or \textbf{Mutation}.

        \label{algline:parent_selection}

        % \STATE Prompt LLM to generate a set of multiple child algorithms $\mathcal{A}_{child}$.
        % \label{algline:generate_child_algorithms}
        \STATE Prompt the LLM to generate a set of child heuristic components $\mathcal{H}_{child}$.
        \label{algline:generate_child_algorithms}

        \STATE Instantiate complete child algorithms $\mathcal{A}_{child} \leftarrow \{B[H] \mid H \in \mathcal{H}_{child}\}$.
        
        \STATE Evaluate $\mathcal{A}_{child}$ on $\mathcal{D}^{train}$.
        \label{algline:evaluate_child_algorithms}
        
        \STATE Update $\mathbb{A}_t$ using the greedy selection and replacement strategies in Eq.~\eqref{eq:greedy_select} and~\eqref{eq:greedy_replace}.
        \label{algline:greedy_select_replace}
        
        \STATE $e_a \leftarrow e_a + |\mathcal{A}_{child}|$.
    \ENDWHILE
\ENDFOR

\RETURN Final portfolio $\mathbb{A}_T$

\end{algorithmic}
\end{algorithm}

\subsection{Potential-Aware Instance Generation}
\label{subsec:llm_data_gen}

\subsubsection{Potential-Aware Instance Evaluation}
A critical challenge in the instance generation process is evaluating the quality of newly generated instances. A straightforward approach would be to incorporate each candidate instance into the training set and re-evolve the portfolio to measure the resulting performance improvement. Traditional methods instead approximate instance quality by absolute hardness, typically measured using the optimality gap to a high-quality reference solution~\cite{Hemert2006, wang2024asp, Branke2011}. However, obtaining such reference solutions is challenging, and absolute hardness does not necessarily indicate that an instance can guide further portfolio improvement. For example, an instance may be challenging for every member algorithm in the current PAP yet offer little room for further performance improvement. PIAC instead uses potential gain $V(x;\mathbb{A}_{t-1})$ as a surrogate for instance-quality evaluation. It measures the current portfolio's improvement potential under controlled perturbations. If perturbing the generated algorithms yields a superior solution, the portfolio retains exploitable room for improvement on that instance, indicating higher estimated instance quality for subsequent evolution.

The operational mechanism of this metric relies on the internal heuristic component of each complete algorithm. Under the fixed backbone $B$, consider a portfolio member $A_j=B[H_j]$, where $H_j$ is the LLM-constructed heuristic function embedded in $B$. For the instance $x\in\Omega$ and the solver state $s$ encountered during solution construction or search, the heuristic component produces a heuristic matrix $M_j=H_j(x,s)$ that guides the subsequent decision. We introduce a \textbf{perturbation operator} $P_{\xi}$ that acts only on this heuristic matrix:
\begin{equation}
    \widetilde{M}_j^{(\xi)}
    = \widetilde{H}_j^{(\xi)}(x,s)
    = P_{\xi}(M_j)
    = M_j \odot
      \max\!\left(
          \varepsilon\mathbf{1},
          \mathbf{1}+E^{(\xi)}
      \right),
    \label{eq:perturb_matrix}
\end{equation}
where $\xi$ indexes an independent realization of the perturbation, $E^{(\xi)}$ is a random matrix with the same dimensions as $M_j$, and $E_{uv}^{(\xi)}\overset{\mathrm{i.i.d.}}{\sim}\mathcal{N}(0,\sigma^2)$. Here, $u$ and $v$ index the entries of $M_j$, $\odot$ denotes element-wise multiplication, $\mathbf{1}$ is the all-ones matrix with the same dimensions as $M_j$, and $\varepsilon>0$ is a small constant that clips the multiplicative factors away from zero. Embedding the perturbed heuristic component into the same backbone yields the complete perturbed algorithm $\widetilde{A}_j^{(\xi)}=B[\widetilde{H}_j^{(\xi)}]$.
 
To illustrate, consider a constructive framework for the Traveling Salesman Problem (TSP) where the heuristic outputs a priority vector $M$ for unvisited nodes. Conventionally, the framework greedily visits the node with the highest priority at each step. Under our formulation, the priority of each node $c$ is perturbed to $\tilde{m}_c = m_c\max(\varepsilon,1+\epsilon_c)$ with $\epsilon_c \sim \mathcal{N}(0, \sigma^2)$. Although the framework strictly retains its standard greedy selection rule ($\tilde{c}^* = \arg\max_{c \in C} \tilde{m}_c$), the injected noise shifts the relative priority rankings and directly alters the chosen decision variables at each step. Consequently, by perturbing the original heuristic strategy, this mechanism effectively simulates a newly constructed prioritization policy that empowers the search process to escape local optima.

Based on this operator, we define the \textbf{potential gain}. Let \(N_p\) denote the total perturbation budget for the current portfolio \(\mathbb A=\{A_1,\ldots,A_K\}\). We distribute this budget as evenly as possible among the \(K\) portfolio members. Specifically, let \(q=\lfloor N_p/K\rfloor\) and \(r_p=N_p\bmod K\). The number of perturbations assigned to \(A_j\) is \(N_{p,j}=q+\mathbb I(j\le r_p)\). Thus, the first \(r_p\) algorithms receive \(q+1\) perturbations each, while the remaining algorithms receive \(q\) perturbations each, ensuring \(\sum_{j=1}^{K}N_{p,j}=N_p\). For each complete algorithm $A_j=B[H_j]$, we perform $N_{p,j}$ independent applications of the perturbation operator $P_{\xi}$ to its heuristic component $H_j$. This produces perturbed heuristic components $\widetilde{H}_j^{(1)},\ldots,\widetilde{H}_j^{(N_{p,j})}$ and the corresponding complete algorithms $\widetilde{A}_j^{(\ell)}=B[\widetilde{H}_j^{(\ell)}]$.
The resulting perturbed algorithm set is $\widetilde{\mathbb{A}}
=\{\widetilde{A}_j^{(\ell)}=B[\widetilde{H}_j^{(\ell)}] \mid j=1,\ldots,K,\; \ell=1,\ldots,N_{p,j}\}$. The potential gain $V(x; \mathbb{A})$ for an instance $x$ is defined as the relative improvement achieved by the best-performing complete perturbed algorithm:
\begin{equation}
    V(x; \mathbb{A}) = \frac{F(\mathbb{A}, x) - F(\widetilde{\mathbb{A}}, x)}{F(\widetilde{\mathbb{A}} \cup \mathbb{A}, x)}.
    \label{eq:perturb_gain}
\end{equation}

If applying slight noise to the scoring matrix yields a significantly better solution than the original deterministic algorithm, it indicates that the portfolio's decision boundary on this instance is suboptimal. This improvement, quantified as the potential gain, allows us to identify instances that expose latent weaknesses in the portfolio, marking them as high-value training data. Because the structure of the scoring matrix and its corresponding perturbation operation depend on the underlying algorithmic architecture, we provide algorithm-specific instantiations of this perturbation mechanism in Section~\ref{sec:instantiations}.

\subsubsection{Instance Generation via Mutator Evolution}
Conventional problem instance generators rely on predefined mutation operators, which often produce homogeneous patterns that fail to adequately expose algorithmic weaknesses~\cite{tang2021ceps}. To overcome this limitation, we conceptualize instance generation as an evolutionary search over the space of mutator programs. We maintain a population of mutators, where the fitness of each mutator $m$ is evaluated by its ability to generate instances with high quality under the current algorithm portfolio $\mathbb{A}_{t-1}$. To ensure a robust evaluation, the mutator $m$ is applied to $N_{exec}$ base instances independently sampled from the initial data distribution $\mathcal{D}_0$, and its overall fitness is defined as the average value across these generated instances:
\begin{equation}
\label{eq:mutator_fitness}
\begin{aligned}
    f(m)
    &= \frac{1}{N_{exec}} 
       \sum_{r=1}^{N_{exec}}
       V\big(m(x_{base}^{(r)}); \mathbb{A}_{t-1}\big), \\
    &\quad \text{where } x_{base}^{(r)} \sim \mathcal{D}_0,
       \quad r=1,\ldots,N_{exec}.
\end{aligned}
\end{equation}
We use potential gain $V(x;\mathbb{A})$, defined in Eq.~\eqref{eq:perturb_gain}, as the surrogate quality score in the mutator fitness function. A higher fitness indicates that the mutator effectively synthesizes instances with higher surrogate quality scores with respect to the current portfolio.

During each evolutionary generation, we sample parent mutators from the current population $\mathcal{M}_{\text{prev}}$ using rank-based stochastic selection, where selection probability is proportional to inverse rank, favoring lower-rank (better-performing) mutators. The LLM is then prompted to perform \textit{crossover} and \textit{mutation} operations to construct a new generation of child mutators $\mathcal{M}_{child}$. The crossover operation recombines parent instance mutators to exploit existing high-quality mutator structures, while the mutation operation enhances exploration by introducing diverse instance mutators (Algorithm~\ref{alg:piac}, lines~\ref{algline:mutator_parent_selection}--\ref{algline:child_mutator_generation}). Full text descriptions of the corresponding prompts can be found in the supplementary material.

After generating $\mathcal{M}_{child}$, we collect all valid mutated instances into a candidate set $\mathcal{X}_{cand}$. To extract the most informative data, we select the $N_{\mathrm{aug}}$ instances that collectively maximize the overall quality:
\begin{equation}
    \mathcal{D}_t = \arg\max_{\substack{\mathcal{S} \subseteq \mathcal{X}_{cand} \\ |\mathcal{S}| = N_{\mathrm{aug}}}} \sum_{x \in \mathcal{S}} V(x; \mathbb{A}_{t-1}).
\end{equation}
Finally, these selected instances are appended to the training dataset ($\mathcal{D}^{train} \leftarrow \mathcal{D}^{train} \cup \mathcal{D}_t$), driving the co-evolution of the subsequent algorithm portfolio (Algorithm~\ref{alg:piac}, lines~\ref{algline:select_informative_instances}--\ref{algline:update_training_dataset}).

\subsection{Automatic Construction of Algorithm Portfolio}
\label{subsec:pap_construct}

Inspired by the EoH-S framework~\cite{liu2025eohs}, we formulate portfolio construction as an evolutionary process, defining the population as the actively maintained algorithm portfolio. The overarching goal is to evolve an algorithm portfolio with strong generalization performance, whose member algorithms exhibit diverse and mutually compensatory behaviors. To achieve this, the framework employs two core mechanisms, namely evolutionary operations for generating novel heuristics and a portfolio maintenance strategy for evaluating and updating the population.

\subsubsection{LLM-Driven Algorithm Evolutionary Operations}
During each iteration, parent algorithms are selected to generate novel child heuristics via two evolutionary operations: the complementary operation and the mutation operation. The complementary operation executes targeted recombination of two mutually compensatory parent heuristics, aiming to synthesize novel algorithms that overcome individual weaknesses and exhibit synergistic search behaviors. In parallel, the mutation operation injects structural variations into existing parent heuristics to continuously foster overall algorithmic diversity.

The execution of the complementary operation begins by pairing mutually complementary parent heuristics. To facilitate this pairing, we employ a parent selection mechanism executed in two stages. In the first stage, the first parent $A_i$ is sampled using rank-based stochastic selection, where the selection probability is proportional to the inverse of the greedy-selection rank $R(A)$ defined via Eq.~\eqref{eq:greedy_select}. In the second stage, the second parent $A_j$ is selected from the remaining candidates based on a complementary gain criterion. Specifically, each candidate is ranked according to its complementary gain with respect to $A_i$, defined as:
\begin{equation}
\begin{split}
    G(A_i, A_j) 
    &= \min\{F(\{A_i\}, \mathcal{D}^{train}), F(\{A_j\}, \mathcal{D}^{train})\} \\
    &\quad - F(\{A_i, A_j\}, \mathcal{D}^{train}),
\end{split}
\label{eq:comp_gain}
\end{equation}
where $F(\cdot)$ denotes portfolio-level performance. A larger $G(A_i, A_j)$ indicates stronger complementarity. 

Following parent selection, we assign specific roles to the parents for the LLM prompt (illustrated in Fig.~\ref{fig:prompt_crossover}). The parent with relatively poorer individual performance is designated as the Reference Algorithm to expose latent algorithmic weaknesses. The better-performing parent serves as the Complementary Algorithm, providing the primary code structure. The LLM then synthesizes a new heuristic by structurally modifying the Complementary Algorithm while explicitly addressing the deficiencies highlighted by the Reference Algorithm. This ensures that the generated algorithms are not mere local refinements but genuinely exhibit complementary search behaviors.
\begin{figure}[!htbp]
\begin{tcolorbox}[
    title=\textbf{Prompt Template: Algorithm Complementary Operation},
    coltitle=white,
    colbacktitle=black!75,
    colback=gray!4!white,
    colframe=black!75,
    boxrule=0.7pt,
    arc=2pt,
    left=6pt, right=6pt, top=8pt, bottom=8pt,
    fontupper=\small\rmfamily\linespread{1.3}\selectfont
]

\textcolor{black!45}{\{user\_generator\}}

Below are two complementary functions. 
They excel in different scenarios or handle different aspects of the problem.

\textbf{[Reference Algorithm Description]} \\
\textcolor{black!45}{\{worse\_algorithm\}}

\textbf{[Reference Algorithm Code]} \\
\textcolor{black!45}{\{worse\_code\}}

\textbf{[Complement Seed Algorithm Description]} \\
\textcolor{black!45}{\{better\_algorithm\}}

\textbf{[Complement Seed Code]} \\
\textcolor{black!45}{\{better\_code\}}

\textbf{[Complementarity Reflection]} \\
\textcolor{black!45}{\{reflection\}}

\textbf{[Complementary Code]} \\
Please write a new function `\textcolor{black!45}{\{func\_name\}\_v2}' which serves as a stronger complementary algorithm to the Reference Algorithm, according to the reflection.

Return `[Algorithm Description]', and one final Python code block.
\end{tcolorbox}
\caption{Prompt for algorithm complementary operation.}

\label{fig:prompt_crossover}
\end{figure}

Additionally, the mutation operation targets individual heuristics to increase exploration. The parent algorithm is sampled with probability inversely proportional to its greedy-selection rank. Taking this selected parent as input, the LLM integrates mutation logic into its structural framework to explore new regions within the algorithmic design space. This process introduces novel algorithmic components and alters existing search trajectories, thereby preventing premature convergence.

To guide evolutionary search, the framework incorporates the reflection mechanism. Rather than evaluating parent heuristics in isolation, this mechanism analyzes sampled historical search trajectories and the current parent. This process generates high-level conceptual hints for refining candidate heuristic structures. The exact prompt formulations driving these components are documented in the Supplement.

\subsubsection{Algorithm Portfolio Construction}
During the initialization phase, we prompt the LLM to generate a
diverse candidate pool $\mathcal{C}$. The candidates are evaluated on the augmented training dataset $\mathcal{D}^{train}$. 
We then construct the portfolio of size $K$ using greedy selection~\cite{liu2025eohs}.
The portfolio is initialized with the
candidate in $\mathcal{C}$ that achieves the best individual
performance on $\mathcal{D}^{train}$. Subsequently, at each step
$h=1, \ldots, K-1$, the candidate with the largest marginal gain is
added:
\begin{equation}
\begin{aligned}
    \Delta_h(A)
    &=
    F\!\left(\mathbb{A}^{(h)},\mathcal{D}^{train}\right)  -
    F\!\left(\mathbb{A}^{(h)}\cup\{A\},
    \mathcal{D}^{train}\right), \\
    A^{(h+1)}
    &=
    \arg\max_{A\in\mathcal{C}\setminus\mathbb{A}^{(h)}}
    \Delta_h(A).
\end{aligned}
\label{eq:greedy_select}
\end{equation}
The selection order defines the \textbf{greedy-selection rank}
$R(A^{(h)})=h$ for $h=1,\ldots,K$, where a lower rank indicates a
larger marginal contribution. During parent selection, portfolio
members are sampled with probabilities proportional to $1/R(A)$,
thereby favoring algorithms with larger marginal contributions.

To manage the integration of newly generated algorithms while strictly maintaining a fixed portfolio capacity $K$, a greedy replacement strategy is implemented. The replacement gain of substituting an existing algorithm $A_i$ with a novel candidate $A_{\mathrm{new}}$ is formulated as:
\begin{equation}
\begin{split}
    \Delta_{\mathrm{rep}}(i; A_{\mathrm{new}}) 
    &= F(\mathbb{A}_t, \mathcal{D}^{train}) \\
    &\quad - F((\mathbb{A}_t \setminus \{A_i\}) \cup \{A_{\mathrm{new}}\}, \mathcal{D}^{train}).
\end{split}
\label{eq:greedy_replace}
\end{equation}
The new candidate replaces the incumbent algorithm that yields the largest replacement gain. The replacement is performed only when this gain is positive, after which the greedy-selection ranks $R(A)$ are recomputed; otherwise, the portfolio remains unchanged. If multiple incumbents yield the same gain, the one with poorer individual performance, measured by a higher mean objective value, is replaced. This update maintains the fixed portfolio size while progressively
improving portfolio-level performance and complementarity (Algorithm~\ref{alg:piac}, Step 2.2).

\section{Algorithm-Specific Instantiations of Perturbation}
\label{sec:instantiations}

The perturbation operator $P_{\xi}$ in Eq.~\eqref{eq:perturb_matrix} acts on the heuristic matrix $M=H(x,s)$, which guides the decisions of the complete algorithm $A=B[H]$. We instantiate this mechanism for three representative backbones: Greedy Constructive, Ant Colony Optimization (ACO), and Guided Local Search (GLS). For each solver setting, the backbone $B$ remains fixed and defines the overall solution procedure, whereas the LLM constructs and evolves only the heuristic component $H$. Perturbing $M$ therefore changes the heuristic guidance used at each solver state without modifying the underlying construction or search process. The following subsections specify $M$ and its role for each backbone.

\subsection{Greedy Constructive Algorithms}
Greedy algorithms iteratively build solutions by selecting candidates $c \in C_t$ based on a heuristic priority matrix $M_t$. Rather than altering the underlying greedy selection rule, our matrix-level perturbation structurally modulates these relative priorities. The perturbed selection becomes:
\begin{equation}
    \widetilde{c}^* = \arg\max_{c \in C_t} \big[P_{\xi}(M_t)\big]_c
\end{equation}
If this perturbation yields a significantly superior final solution, it implies the instance is highly sensitive to the heuristic's local scoring logic. Such instances are highly valuable for evaluation, as they explicitly expose critical vulnerabilities in standard greedy decision-making.

\subsection{Ant Colony Optimization (ACO)}
In ACO~\cite{ye2023deepaco}, the search is guided by both dynamic pheromone trails $\tau_{ij}$ and a static heuristic information matrix $M$. Under our framework, pheromone updates proceed conventionally based on search history, but the heuristic matrix is replaced by its perturbed counterpart $P_{\xi}(M)$. The perturbed candidate-edge weight is computed as:
\begin{equation}
    \widetilde{W}_{ij} = \tau_{ij}^{\alpha} [P_{\xi}(M)]_{ij}^{\beta}
\end{equation}
where $\alpha$ and $\beta$ balance the pheromone and heuristic influences. By altering the ants' sampling distribution, we can evaluate the instance's underlying difficulty. If the perturbed matrix enables ACO to escape local optima and find better solutions, the instance serves as a valuable stress test, revealing cases where the original heuristic inappropriately biased the search trajectory.

\subsection{Guided Local Search (GLS)}
In GLS~\cite{arnold2019kgls}, the matrix $M$ governs edge-level penalty selection to help the search escape local minima. Applying $P_{\xi}$ directly reshapes the utility landscape dictating this penalization. The target edge $(i^*,j^*)$ in the current solution $S$ is selected via:
\begin{equation}
    (i^*,j^*) = \arg\max_{(i,j)\in E(S)} \frac{[P_{\xi}(M)]_{ij}}{1+p_{ij}}
\end{equation}
where $p_{ij}$ is the accumulated penalty. Following this selection, GLS updates the penalty-adjusted cost matrix (i.e., $W_{ij} = c_{ij} + k_{GLS} p_{ij}$) conventionally.

A high potential gain from this perturbation reveals that $M$ originally formed a ``utility trap'' by overvaluing suboptimal edges for this specific instance. Incorporating these highly sensitive, vulnerable instances into training forces the evolutionary search to circumvent such traps, ultimately driving the evolution of a more generalizable search policy.

\section{Experiments}
\label{sec:experiments}
\begin{table}[t]
\centering
\caption{Hyperparameter settings of PIAC.}
\label{tab:hyperparameters}
\begin{tabular}{llc}
\toprule
\textbf{Parameter} & \textbf{Symbol} & \textbf{Value} \\
\midrule
\multicolumn{3}{l}{\textbf{Framework}} \\
Max iterations & $T$ & 4 \\
\midrule
\multicolumn{3}{l}{\textbf{Algorithm evolution}} \\
Portfolio size & $K$ & 5 \\
\# Algorithm evals/iter & $FE_a$ & 100 \\
\midrule
\multicolumn{3}{l}{\textbf{Data evolution}} \\
Initial dataset size & $|\mathcal{D}_0|$ & 8 \\
\# Mutator evals/iter & $FE_m$ & 30 \\
\# Executions per mutator & $N_{\mathrm{exec}}$ & 4 \\
\# Augmented instances/iter & $N_{\mathrm{aug}}$ & 8 \\
\# Noise perturbations & $N_p$ & 64 \\
Noise strength & $\sigma$ & 0.01 \\
\bottomrule
\end{tabular}
\end{table}
We conduct extensive experiments to evaluate the proposed method on two representative combinatorial optimization problems: the Traveling Salesman Problem (TSP) and the Capacitated Vehicle Routing Problem (CVRP). Our experiments are designed to answer four research questions (\emph{RQs}):

\begin{list}{}{
    \setlength{\leftmargin}{0.13\linewidth}
    \setlength{\labelwidth}{0.10\linewidth}
    \setlength{\labelsep}{0.03\linewidth}
    \setlength{\itemsep}{0.15em}
    \setlength{\parsep}{0pt}
    \setlength{\topsep}{0.2em}
}
    \item[\emph{RQ1}:] Does the proposed framework construct an algorithm portfolio that generalizes across different data distributions?

    \item[\emph{RQ2}:] Do the two proposed components effectively contribute to the overall performance of the framework?

    \item[\emph{RQ3}:] What is the relationship between the potential gain and the optimality gap?
    
    \item[\emph{RQ4}:] Do the newly generated valuable instances continuously improve the performance of the algorithm portfolio?
\end{list}

\subsection{Experimental Setup}

\begin{table*}[htbp]
\centering

\caption{
Performance of various methods on constructive heuristic design for synthetic TSP and CVRP across six instance distributions. 
Values on the left and right under each distribution represent the objective value (Obj) and optimality gap (Gap), respectively (lower is better). 
Bold values indicate the best result and any results not significantly different from it (paired Wilcoxon signed-rank test with Holm correction, $\alpha=0.05$).
}
\label{tab:main_synthetic_constructive}

\begingroup
\scriptsize
\setlength{\tabcolsep}{2pt}
\renewcommand{\arraystretch}{0.82}

\resizebox{0.88\textwidth}{!}{%
\begin{tabular}{lccccccccccccc}

\toprule

\multicolumn{14}{c}{\textbf{TSP Constructive Heuristic}} \\
\midrule

Method
& \multicolumn{2}{c}{Rue}
& \multicolumn{2}{c}{Explosion}
& \multicolumn{2}{c}{Implosion}
& \multicolumn{2}{c}{Expansion}
& \multicolumn{2}{c}{Cluster}
& \multicolumn{2}{c}{Grid}
& \multicolumn{1}{c}{Avg} \\
\midrule

FunSearch
& 12.91 & 20.75\%
& 10.27 & 22.44\%
& 10.83 & 24.49\%
& 10.59 & 21.99\%
& 9.71  & 23.15\%
& 13.11 & 18.14\%
& 21.83\% \\

EoH
& 12.49 & 16.84\%
& 10.05 & 19.83\%
& 10.50 & 20.78\%
& 10.26 & 18.26\%
& 9.47  & 20.08\%
& 12.75 & 14.83\%
& 18.44\% \\

ReEvo
& 12.73 & 19.05\%
& 10.15 & 21.04\%
& 10.63 & 22.19\%
& 10.41 & 19.92\%
& 9.49  & 20.21\%
& 12.98 & 16.97\%
& 19.90\% \\

MCTS-AHD
& 12.97 & 21.31\%
& 10.43 & 24.38\%
& 10.81 & 24.27\%
& 10.60 & 22.12\%
& 9.76  & 23.71\%
& 13.21 & 19.05\%
& 22.47\% \\

EoH-S
& 12.25 & 14.54\%
& 9.73  & 16.00\%
& 10.09 & 15.99\%
& 9.97  & 14.93\%
& 9.07  & 14.88\%
& 12.51 & 12.69\%
& 14.83\% \\

PIAC (RND)
& 12.05 & 12.69\%
& 9.67  & 15.32\%
& 9.92  & 14.07\%
& 9.85  & 13.48\%
& 8.98  & 13.82\%
& 12.37 & 11.45\%
& 13.47\% \\

PIAC (GAP)
& 12.01 & 12.30\%
& 9.66  & 15.16\%
& 9.90  & 13.82\%
& 9.80  & 12.99\%
& 8.94  & 13.29\%
& 12.34 & 11.14\%
& 13.12\% \\

PIAC
& 11.93 & \textbf{11.54\%}
& 9.49  & \textbf{13.23\%}
& 9.80  & \textbf{12.68\%}
& 9.70  & \textbf{11.84\%}
& 8.85  & \textbf{12.23\%}
& 12.20 & \textbf{9.90\%}
& \textbf{11.90\%} \\

\midrule

\multicolumn{14}{c}{\textbf{CVRP Constructive Heuristic}} \\
\midrule

Method
& \multicolumn{2}{c}{Rue}
& \multicolumn{2}{c}{Explosion}
& \multicolumn{2}{c}{Implosion}
& \multicolumn{2}{c}{Expansion}
& \multicolumn{2}{c}{Cluster}
& \multicolumn{2}{c}{Grid}
& \multicolumn{1}{c}{Avg} \\
\midrule

FunSearch
& 34.20 & 24.45\%
& 32.60 & 22.41\%
& 31.48 & 23.62\%
& 30.54 & 21.87\%
& 31.45  & 19.89\%
& 34.31 & 25.74\%
& 23.00\% \\

EoH
& 34.37 & 25.00\%
& 32.47 & 21.94\%
& 31.15 & 22.38\%
& 30.63 & 22.23\%
& 31.20 & 18.92\%
& 34.54 & 26.53\%
& 22.83\% \\

ReEvo
& 34.57 & 25.77\%
& 33.05 & 24.21\%
& 31.66 & 24.33\%
& 31.14 & 24.30\%
& 31.65 & 20.59\%
& 34.47 & 26.20\%
& 24.23\% \\

MCTS-AHD
& 34.77 & 26.54\%
& 33.06 & 24.44\%
& 31.78 & 24.96\%
& 31.18 & 24.53\%
& 31.84 & 21.52\%
& 34.67 & 27.08\%
& 24.85\% \\

EoH-S
& 33.67 & 22.55\%
& 31.94 & 20.22\%
& 30.60 & 20.32\%
& 30.24 & 20.74\%
& 30.58 & 16.69\%
& 33.58 & 23.06\%
& 20.60\% \\

PIAC (RND)
& 33.34 & 21.24\%
& 31.43 & 17.99\%
& 30.23 & 18.78\%
& 29.69 & 18.47\%
& 30.49 & 16.17\%
& 33.33 & \textbf{22.09}\%
& 19.12\% \\

PIAC (GAP)
& 33.19 & \textbf{20.78\%}
& 31.21 & \textbf{17.16}\%
& 30.05 & \textbf{18.07}\%
& 29.52 & \textbf{17.79}\%
& 30.36 & \textbf{15.72}\%
& 33.25 & \textbf{21.81}\%
& 18.56\% \\

PIAC
& 33.07 & \textbf{20.36\%}
& 31.10 & \textbf{16.77\%}
& 29.97 & \textbf{17.79\%}
& 29.46 & \textbf{17.59\%}
& 30.30 & \textbf{15.48\%}
& 33.23 & \textbf{21.76\%}
& \textbf{18.29\%} \\

\bottomrule
\end{tabular}%
}

\endgroup
\end{table*}
\textbf{Datasets.} For each problem class, the training instances are generated under the random-distribution setting adopted in ReEvo.
The test instances follow Bossek et al.~\cite{Bossek2019syninst, Zhou2025syninstsurvey}, where structured locations are generated from random uniform Euclidean (Rue) point clouds by simulating five spatial patterns: explosion, implosion, cluster, expansion, and grid.
The same location distributions are used for both TSP and CVRP.
For all synthetic instances, the problem size is fixed to $n=200$.
In this work, we use 8 training instances for each problem class and generate 100 test instances for each spatial pattern.

\textbf{Evaluation Metric.} We report the optimality gap to evaluate the solution quality of each method.
For each instance $x$, the gap is computed with respect to a reference objective value.
Specifically, we use LKH~\cite{Helsgaun2000lkh} for TSP instances and HGS~\cite{Vidal2022hgs} for CVRP instances to obtain the reference objective value.
Given the objective value \(F(\mathbb A,x)\) obtained by the algorithm portfolio \(\mathbb A\) on instance \(x\) and the reference objective value $f_{\mathrm{ref}}(x)$, the optimality gap is defined as:
\begin{equation}
    \mathrm{Gap}(\mathbb A, x) =
    \frac{F(\mathbb A, x) - f_{\mathrm{ref}}(x)}
    {f_{\mathrm{ref}}(x)} \times 100\%.
\end{equation}
Lower gap values indicate better performance.

\textbf{Compared Methods.} We compare the proposed method with representative LLM-based baselines, specifically FunSearch~\cite{romera2023funsearch}, EoH~\cite{liu2024eoh}, ReEvo~\cite{ye2024reevo}, MCTS-AHD~\cite{zheng2025mcts}, and EoH-S~\cite{liu2025eohs}.
Direct comparison with traditional frameworks like CEPS remains infeasible because they optimize continuous parameter spaces rather than discrete heuristics. To indirectly evaluate our instance evolution against CEPS strategies, we introduce two strong baselines within the algorithm space, as detailed below: 
\begin{itemize}
    \item \textbf{PIAC (RND):} This variant replaces the LLM-based mutator with a fixed random (RND) perturbation operator, while retaining potential gain as the instance evaluation metric.
    \item \textbf{PIAC (GAP):} This variant retains the LLM-based mutator but replaces potential gain with the \textit{Optimality Gap} (evaluated via strong solvers) as the instance evaluation metric.
\end{itemize}
Since the base training dataset remains identical during the initial 100 evaluations, we directly reuse the algorithm portfolio evolved by the full PIAC at $FE_a = 100$ to ensure an identical starting point for all comparisons.

\textbf{Implementation Details.} Unless otherwise specified, we employ DeepSeek-V3.2~\cite{deepseekv3} as the backbone LLM for both algorithm and generator queries. For the algorithm evolution process, the total number of algorithm evaluations is configured to $T \cdot FE_a = 400$. Specifically, after every $FE_a = 100$ algorithm evaluations, the data evolution process conducts 30 instance mutator evaluations. From these generated candidates, 8 newly constructed problem instances are selected and incorporated into the training set for algorithm evaluation. The primary hyperparameters utilized in our framework are summarized in Table~\ref{tab:hyperparameters}. To ensure a fair comparison, these same hyperparameter settings are applied across all the baseline methods introduced above. Each experiment is independently repeated 3 times, and we report the average performance across these runs.

\subsection{Overall Results}

\begin{table*}[htbp]
\centering
\caption{
Performance comparison of different methods for designing heuristics within two TSP backbones: ACO and GLS, across six instance distributions. 
Values on the left and right under each distribution represent the objective value (Obj) and optimality gap (Gap), respectively (lower is better). 
Bold values indicate the best result and any results not significantly different from it (paired Wilcoxon signed-rank test with Holm correction, $\alpha=0.05$).
}
\label{tab:main_synthetic_frameworks}

\begingroup
\scriptsize
\setlength{\tabcolsep}{2pt}
\renewcommand{\arraystretch}{0.82}

\resizebox{0.88\textwidth}{!}{%
\begin{tabular}{lccccccccccccc}
\toprule

\multicolumn{14}{c}{\textbf{ACO}} \\
\midrule

Method
& \multicolumn{2}{c}{Rue}
& \multicolumn{2}{c}{Explosion}
& \multicolumn{2}{c}{Implosion}
& \multicolumn{2}{c}{Expansion}
& \multicolumn{2}{c}{Cluster}
& \multicolumn{2}{c}{Grid}
& \multicolumn{1}{c}{Avg} \\
\midrule

FunSearch
& 11.88 & 11.06\%
& 9.45  & 12.55\%
& 9.63  & 10.68\%
& 9.66  & 11.29\%
& 8.71  & 10.28\%
& 12.24 & 10.26\%
& 11.02\% \\

EoH
& 11.96 & 11.81\%
& 9.49  & 13.12\%
& 9.71  & 11.61\%
& 9.69  & 11.71\%
& 8.77  & 11.11\%
& 12.30 & 10.81\%
& 11.70\% \\

ReEvo
& 12.20 & 14.06\%
& 9.70  & 15.64\%
& 10.00 & 15.01\%
& 9.89  & 14.07\%
& 9.07  & 15.04\%
& 12.50 & 12.66\%
& 14.41\% \\

MCTS-AHD
& 11.98 & 12.02\%
& 9.52  & 13.39\%
& 9.75  & 12.10\%
& 9.71  & 11.97\%
& 8.78  & 11.24\%
& 12.33 & 11.06\%
& 11.96\% \\

EoH-S
& 11.78 & 10.16\%
& 9.36  & 11.49\%
& 9.58  & 10.12\%
& 9.58  & 10.40\%
& 8.64  & 9.49\%
& 12.16 & 9.51\%
& 10.20\% \\

PIAC (RND)
& 11.60 & 8.49\%
& 9.19  & 9.53\%
& 9.45  & 8.61\%
& 9.42  & 8.61\%
& 8.56  & 8.52\%
& 11.97 & 7.89\%
& 8.61\% \\

PIAC (GAP)
& 11.66 & 9.01\%
& 9.22  & 9.93\%
& 9.47  & 8.90\%
& 9.46  & 9.05\%
& 8.55  & 8.35\%
& 12.01 & 8.23\%
& 8.91\% \\

PIAC
& 11.56 & \textbf{8.06\%}
& 9.16  & \textbf{9.11\%}
& 9.37  & \textbf{7.67\%}
& 9.40  & \textbf{8.32\%}
& 8.46  & \textbf{7.22\%}
& 11.94 & \textbf{7.61\%}
& \textbf{8.00\%} \\

\midrule

\multicolumn{14}{c}{\textbf{GLS}} \\
\midrule

Method
& \multicolumn{2}{c}{Rue}
& \multicolumn{2}{c}{Explosion}
& \multicolumn{2}{c}{Implosion}
& \multicolumn{2}{c}{Expansion}
& \multicolumn{2}{c}{Cluster}
& \multicolumn{2}{c}{Grid}
& \multicolumn{1}{c}{Avg} \\
\midrule

FunSearch
& 10.714 & 0.175\%
& 8.402  & 0.090\%
& 8.714  & 0.173\%
& 8.698  & 0.247\%
& 7.899  & 0.128\%
& 11.124 & 0.209\%
& 0.170\% \\

EoH
& 10.715 & 0.183\%
& 8.405  & 0.120\%
& 8.716  & 0.202\%
& 8.700  & 0.266\%
& 7.903  & 0.172\%
& 11.128 & 0.241\%
& 0.197\% \\

ReEvo
& 10.716 & 0.194\%
& 8.405  & 0.124\%
& 8.714  & 0.182\%
& 8.699  & 0.261\%
& 7.902  & 0.170\%
& 11.123 & 0.196\%
& 0.187\% \\

MCTS-AHD
& 10.715 & 0.185\%
& 8.404  & 0.114\%
& 8.712  & 0.157\%
& 8.696  & 0.228\%
& 7.903  & 0.178\%
& 11.126 & 0.224\%
& 0.181\% \\

EoH-S
& 10.701 & 0.057\%
& 8.397  & \textbf{0.024\%}
& 8.703  & \textbf{0.055\%}
& 8.683  & \textbf{0.074\%}
& 7.893  & 0.051\%
& 11.110 & 0.077\%
& 0.056\% \\

PIAC (RND)
& 10.699 & \textbf{0.041}\%
& 8.397  & \textbf{0.026}\%
& 8.703  & \textbf{0.049}\%
& 8.683  & \textbf{0.072\%}
& 7.893  & 0.054\%
& 11.107 & \textbf{0.055\%}
& \textbf{0.050\%} \\

PIAC (GAP)
& 10.699 & \textbf{0.042\%}
& 8.396  & \textbf{0.022\%}
& 8.702  & \textbf{0.043\%}
& 8.682  & \textbf{0.065\%}
& 7.891  & \textbf{0.029\%}
& 11.108 & \textbf{0.060\%}
& \textbf{0.044\%} \\

PIAC
& 10.699 & \textbf{0.040\%}
& 8.396  & \textbf{0.018\%}
& 8.702  & \textbf{0.047\%}
& 8.685  & \textbf{0.098\%}
& 7.892  & \textbf{0.036\%}
& 11.108 & \textbf{0.063\%}
& \textbf{0.050\%} \\

\bottomrule
\end{tabular}%
}

\endgroup
\end{table*}
\subsubsection{Performance on Diverse Problem Distributions}

Table~\ref{tab:main_synthetic_constructive} reports the performance of all evaluated methods using the Greedy Construction backbone on synthetic TSP and CVRP instances. To evaluate generalization capabilities, all methods are trained exclusively on the rue training dataset and subsequently tested across six distinct distributions, comprising the in-distribution rue instances and five additional unseen distributions. For constructive heuristics, PIAC consistently achieves the lowest optimality gaps, demonstrating strong generalization well beyond the training set. Notably, PIAC significantly outperforms single-algorithm baselines, including EoH, ReEvo, and MCTS. For instance, on the constructive TSP, PIAC achieves an average gap of $11.90\%$, substantially lower than those of EoH ($18.44\%$), ReEvo ($19.90\%$), and MCTS ($22.47\%$). This confirms that a complementary portfolio of heuristics is inherently more effective than discovering a single best algorithm. Furthermore, compared to EoH-S, PIAC achieves a $19.76\%$ relative reduction in the average gap on TSP (from $14.83\%$ to $11.90\%$) and an $11.21\%$ relative reduction on CVRP (from $20.60\%$ to $18.29\%$). While EoH-S also constructs an algorithm portfolio using fixed instances, PIAC dynamically co-evolves the complementary algorithms alongside valuable synthesized data. These outcomes directly answer \emph{RQ1} by demonstrating that our co-evolutionary framework constructs an algorithm portfolio capable of robustly generalizing across diverse data distributions, rather than simply overfitting to the initial rue training set.

This robust generalization extends across diverse algorithms, as shown in Table~\ref{tab:main_synthetic_frameworks}. On TSP ACO, PIAC reduces the average gap from $10.20\%$ (EoH-S) to $8.00\%$. For TSP GLS, where baseline methods already achieve near-zero gaps and inherently limit further improvements, PIAC continues to outperform EoH-S, successfully reducing the average gap from $0.056\%$ to $0.050\%$.

To further validate generalization capabilities, we benchmark the portfolios on standard public datasets, specifically TSPLib~\cite{Reinelt1991tsplib} for TSP and CVRPLib~\cite{Uchoa2017cvrplib} for CVRP. As shown in Table~\ref{tab:public_benchmarks}, PIAC attains the lowest optimality gap on TSPLib, decreasing it from $13.67\%$ (EoH-S) to $11.79\%$. On CVRP, PIAC secures the top rank across 8 of the 10 CVRPLib subsets and places second on the remaining two, dropping the overall average gap to $24.53\%$ compared to the $29.88\%$ achieved by EoH-S. Collectively, these findings directly answer \emph{RQ1} by demonstrating that the proposed co-evolutionary framework successfully constructs algorithm portfolios capable of robust generalization across diverse data distributions.

\begin{table}[t]
\centering
\caption{
Performance on public benchmark instances.
Lower values are better.
Bold values indicate the best result and any results not significantly different from it (paired Wilcoxon signed-rank test with Holm correction, $\alpha=0.05$).
}
\label{tab:public_benchmarks}
\scriptsize
\setlength{\tabcolsep}{8pt}
\resizebox{\linewidth}{!}{
\begin{tabular}{lccccc}
\toprule
\textbf{Benchmarks}
& \textbf{ReEvo}
& \textbf{EoH-S}
& \textbf{\makecell{PIAC\\(RND)}}
& \textbf{\makecell{PIAC\\(GAP)}}
& \textbf{\makecell{PIAC}} \\
\midrule

TSPLib
& 19.59\%
& 13.67\%
& 13.28\%
& 12.92\%
& \textbf{11.79\%} \\

\midrule

CVRPLib A
& 29.60\%
& 25.74\%
& 25.36\%
& \textbf{24.02\%}
& \textbf{22.97\%} \\

CVRPLib B
& 32.37\%
& 24.75\%
& \textbf{17.90\%}
& \textbf{18.10\%}
& \textbf{16.97\%} \\

CVRPLib CMT
& 39.17\%
& \textbf{33.34\%}
& 34.65\%
& \textbf{32.15\%}
& \textbf{32.95\%} \\

CVRPLib F
& \textbf{50.70\%}
& \textbf{44.43\%}
& \textbf{44.35\%}
& \textbf{36.85\%}
& \textbf{33.97\%} \\

CVRPLib Golden
& 26.82\%
& \textbf{22.86\%}
& \textbf{22.22\%}
& 22.69\%
& \textbf{20.58\%} \\

CVRPLib Li
& \textbf{27.89\%}
& \textbf{25.49\%}
& \textbf{21.42\%}
& \textbf{20.73\%}
& \textbf{18.26\%} \\

CVRPLib M
& \textbf{38.65\%}
& \textbf{36.17\%}
& \textbf{34.31\%}
& \textbf{33.21\%}
& \textbf{33.75\%} \\

CVRPLib P
& \textbf{25.21\%}
& \textbf{21.77\%}
& \textbf{21.39\%}
& \textbf{20.71\%}
& \textbf{20.22\%} \\

CVRPLib tai
& 51.52\%
& 43.64\%
& 33.57\%
& \textbf{30.93\%}
& \textbf{28.53\%} \\

CVRPLib X
& 24.05\%
& 20.65\%
& 18.04\%
& 17.76\%
& \textbf{17.12\%} \\

\bottomrule
\end{tabular}
}
\end{table}

\subsubsection{Effectiveness of the Proposed Components}
To answer \emph{RQ2}, we conduct an ablation study to validate the core mechanisms of PIAC. To isolate the contributions of our co-evolutionary design, we evaluate the full framework against two strong internal variants. 

First, we assess the impact of the LLM-driven instance generation by introducing PIAC (RND). This variant replaces the LLM-constructed mutator with a fixed random perturbation operator while retaining the proposed potential gain for instance evaluation. The evaluation reveals that the full PIAC outperforms the PIAC (RND) baseline, reducing the average optimality gap from $13.47\%$ to $11.90\%$ on the constructive TSP and from $19.12\%$ to $18.29\%$ on the constructive CVRP. This robust improvement demonstrates that LLM-evolved mutators successfully explore a broader data space to effectively enhance generalization. 

Second, we evaluate the effectiveness of the instance evaluation metric by comparing our full framework against PIAC (GAP). This variant retains the LLM mutator but relies on the traditional optimality gap evaluated via strong external solvers to assess generated instances. The full framework improves the average gap from $13.12\%$ to $11.90\%$ on the TSP Constructive Heuristic and from $8.91\%$ to $8.00\%$ on TSP ACO without requiring reference solvers. This advantage arises because potential gain exposes suboptimal decision boundaries, precisely targeting high-value instances that reveal latent algorithmic weaknesses rather than simply selecting universally hard problems.

Collectively, these findings directly answer \emph{RQ2} by confirming that both the LLM-driven mutator evolution and the potential-aware metric contribute to constructing robust algorithm portfolios.

\subsubsection{Time Cost Analysis}
To evaluate the computational efficiency of the proposed metric, we compare the time cost of potential gain against the traditional Opt Gap. We evaluate 120 Rue problem instances using a reference algorithm portfolio constructed after 100 function evaluations. The optimal objectives for Opt Gap are computed using default configurations of LKH for TSP and HGS for CVRP. Both metrics are computed using 64 parallel processes.

Table~\ref{tab:cost_analysis} details the evaluation results. Regarding the time cost, the potential gain demonstrates highly competitive efficiency. With the sole exception of TSP Constructive, where Opt Gap holds a minor computational edge, the proposed metric matches or significantly accelerates the evaluation process across all other paradigms. Most notably on CVRP Constructive instances, potential gain requires only 9.46 seconds, which is substantially faster than the 46.35 seconds consumed by Opt Gap. For both TSP ACO and TSP GLS, the execution times of the two metrics are strictly competitive and closely matched.

Unlike Opt Gap, whose computational overhead fluctuates with external reference solvers, potential gain avoids these dependencies. By relying on controlled perturbations, its evaluation time is bounded by the execution time of the algorithm itself. Empirical results show that the evaluation cost of \textit{potential gain} remains relatively stable.
\begin{table}[htbp]
\centering
\caption{
Cost analysis of Potential Gain vs Opt Gap. 
The reported values are computational time in seconds.
}
\label{tab:cost_analysis}
\begin{tabular}{lcc}
\toprule
\textbf{Problem}
& \textbf{Potential Gain / s}
& \textbf{Opt Gap / s} \\
\midrule

TSP Constructive
& 8.91
& \textbf{3.59} \\

CVRP Constructive
& \textbf{9.46}
& 46.35 \\

TSP ACO
& \textbf{16.63}
& 17.46 \\

TSP GLS
& 8.28
& \textbf{6.37} \\

\bottomrule
\end{tabular}
\end{table}

\begin{figure*}[t]
    \centering
    \begin{minipage}[t]{0.35\textwidth}
        \centering
        \vspace{0pt}
        \includegraphics[width=\linewidth]{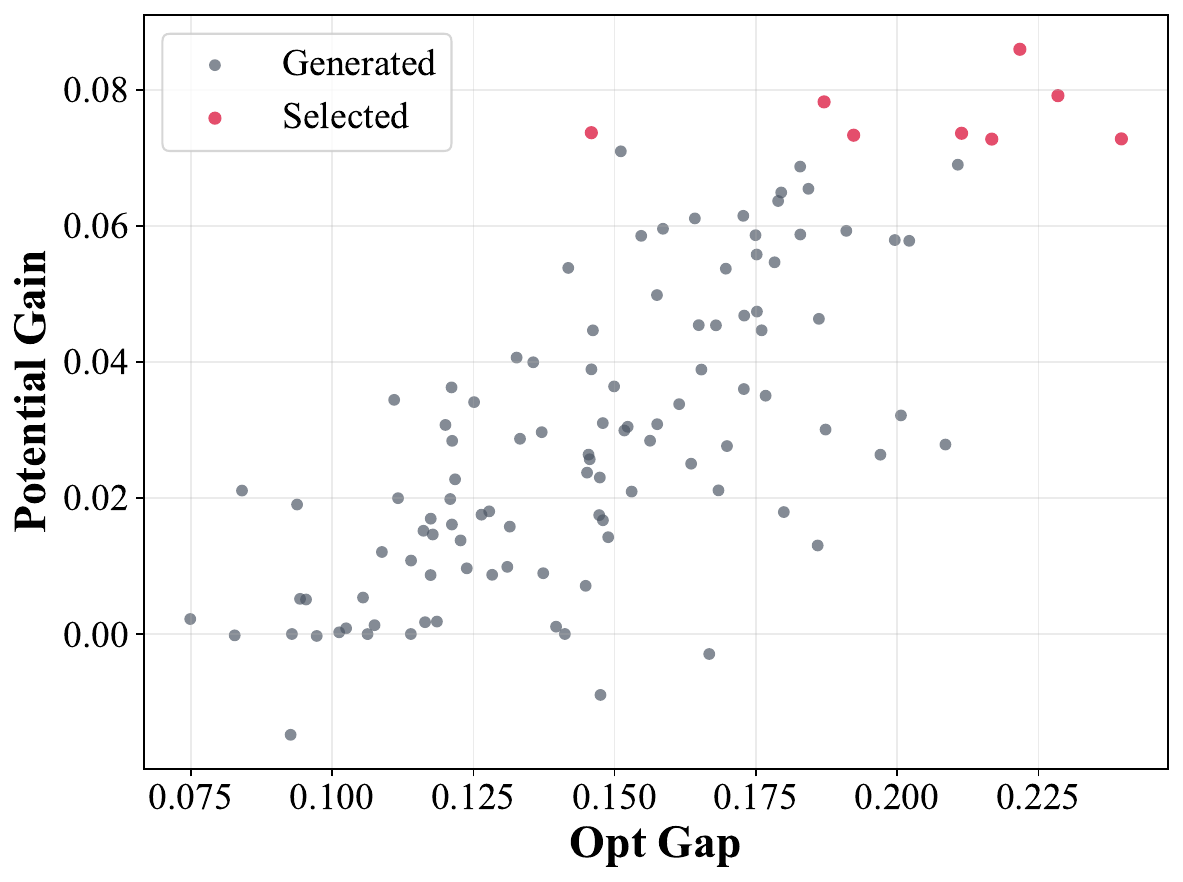}
        \\[0.4em]
        {\small \textbf{(a)} TSP}
    \end{minipage}
    \hspace{0.08\textwidth}
    \begin{minipage}[t]{0.35\textwidth}
        \centering
        \vspace{0pt}
        \includegraphics[width=\linewidth]{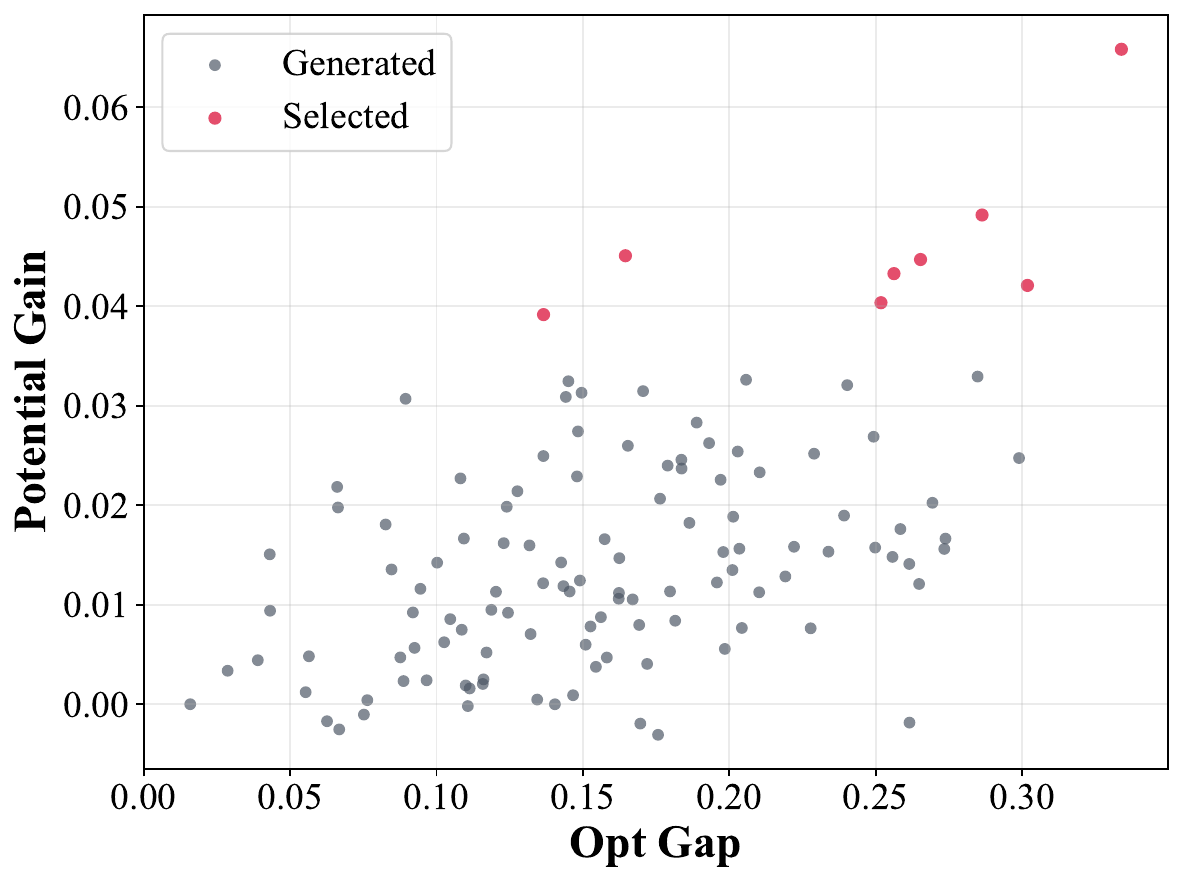}
        \\[0.4em]
        {\small \textbf{(b)} CVRP}
    \end{minipage}

    \caption{
    Relationship between the potential gain and the optimality gap on generated instances.
    }
    \label{fig:noise_gap_vs_opt_gap}
\end{figure*}

\subsection{Relationship Between Potential Gain and Optimality Gap}

To address \emph{RQ3}, we investigate whether the proposed potential gain positively correlates with the hardness metric. To ensure a fair evaluation, we freeze the algorithm portfolio (PAP) after $FE_a = 100$ evaluations for each problem domain (TSP constructive and CVRP constructive heuristics). We then use this static PAP to evaluate all problem instances generated by the LLM-evolved mutators. This setup ensures that instance hardness is measured against a consistent algorithmic baseline, isolating the evaluation from the dynamics of a continuously evolving portfolio.

For each training instance $i$, let $v_i$ denote its potential gain and $g_i$ denote its optimality gap, both measured using the static PAP. We quantify their statistical relationship via the Pearson correlation coefficient:
\begin{equation}
    \rho_{v,g} =
    \frac{
    \sum_{i=1}^{N}(v_i-\bar{v})(g_i-\bar{g})
    }{
    \sqrt{\sum_{i=1}^{N}(v_i-\bar{v})^2}
    \sqrt{\sum_{i=1}^{N}(g_i-\bar{g})^2}
    },
\end{equation}
where $N$ is the total number of evaluated instances, and $\bar{v}$ and $\bar{g}$ are the sample means of the potential gains and optimality gaps, respectively.

As illustrated in Figure~\ref{fig:noise_gap_vs_opt_gap}, potential gain is positively associated with the optimality gap. The Pearson correlation coefficients are $0.7445$ for TSP and $0.5230$ for CVRP, indicating strong and moderate positive correlations, respectively. These results suggest that harder instances tend to exhibit greater improvement potential. However, some hard instances show only limited improvement after perturbation. The potential gain metric therefore favors instances with greater potential for improvement, rather than simply those with large optimality gaps. As a result, many selected instances have both high potential gains and large optimality gaps, without requiring reference solutions during instance evaluation.
These results answer \emph{RQ3} by showing that potential gain is positively correlated with instance hardness while capturing a distinct, improvement-oriented signal rather than merely reproducing the optimality gap.

\begin{figure*}[t]
    \centering
    \begin{minipage}[t]{0.24\textwidth}
        \centering
        \includegraphics[width=\linewidth]{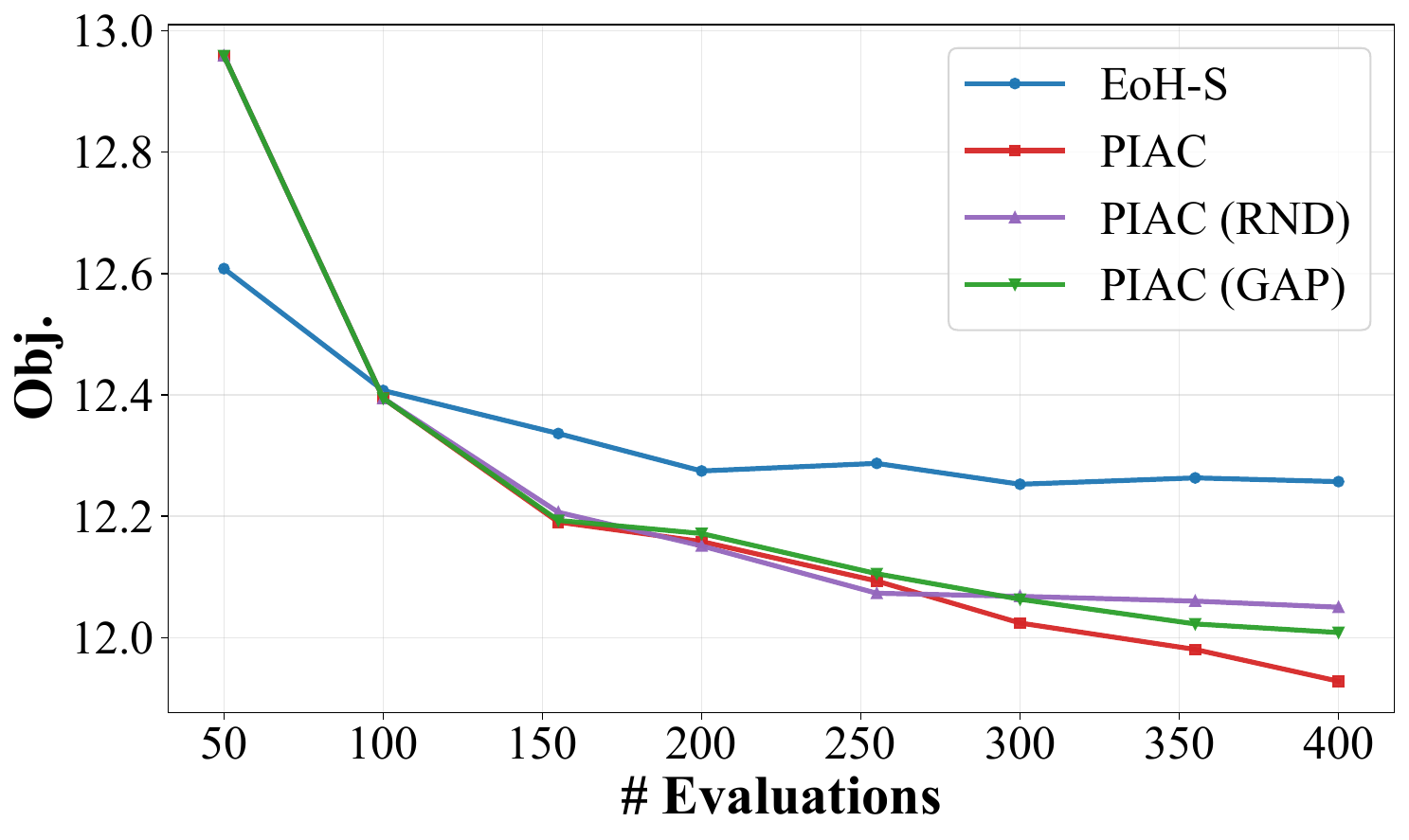}\\[-1mm]
        {\small (a) TSP-Rue}
    \end{minipage}
    \hfill
    \begin{minipage}[t]{0.24\textwidth}
        \centering
        \includegraphics[width=\linewidth]{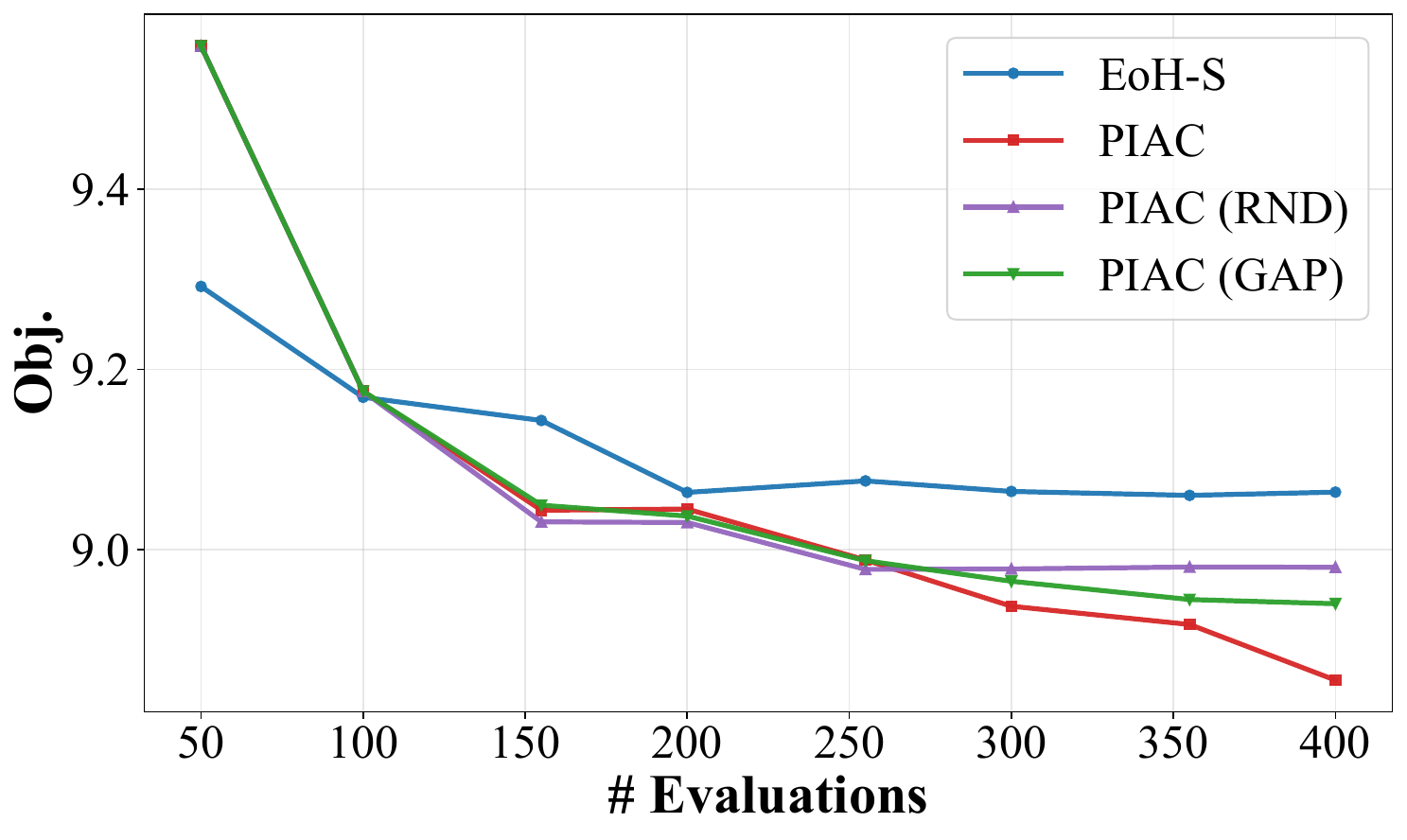}\\[-1mm]
        {\small (b) TSP-Cluster}
    \end{minipage}
    \hfill
    \begin{minipage}[t]{0.24\textwidth}
        \centering
        \includegraphics[width=\linewidth]{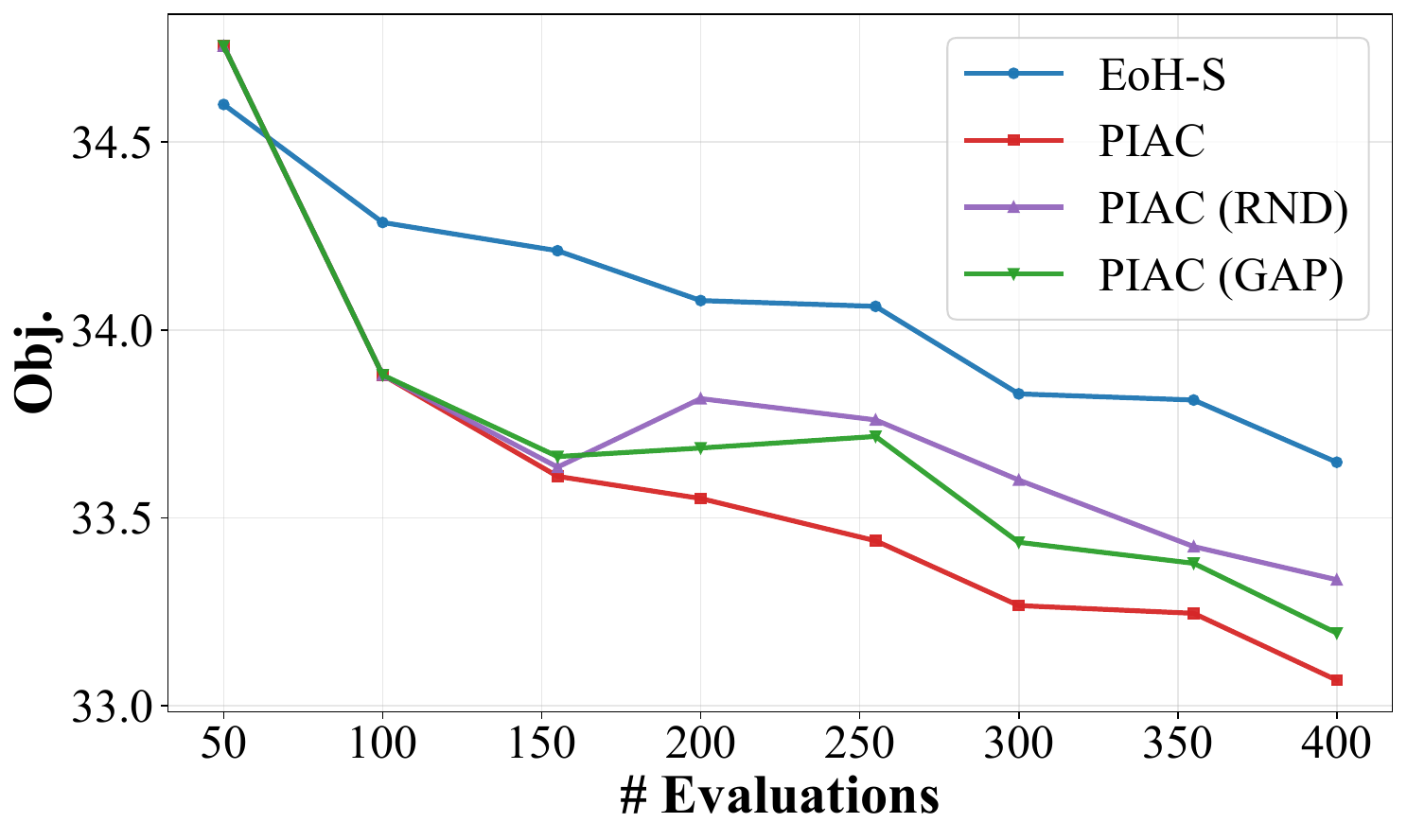}\\[-1mm]
        {\small (c) CVRP-Rue}
    \end{minipage}
    \hfill
    \begin{minipage}[t]{0.24\textwidth}
        \centering
        \includegraphics[width=\linewidth]{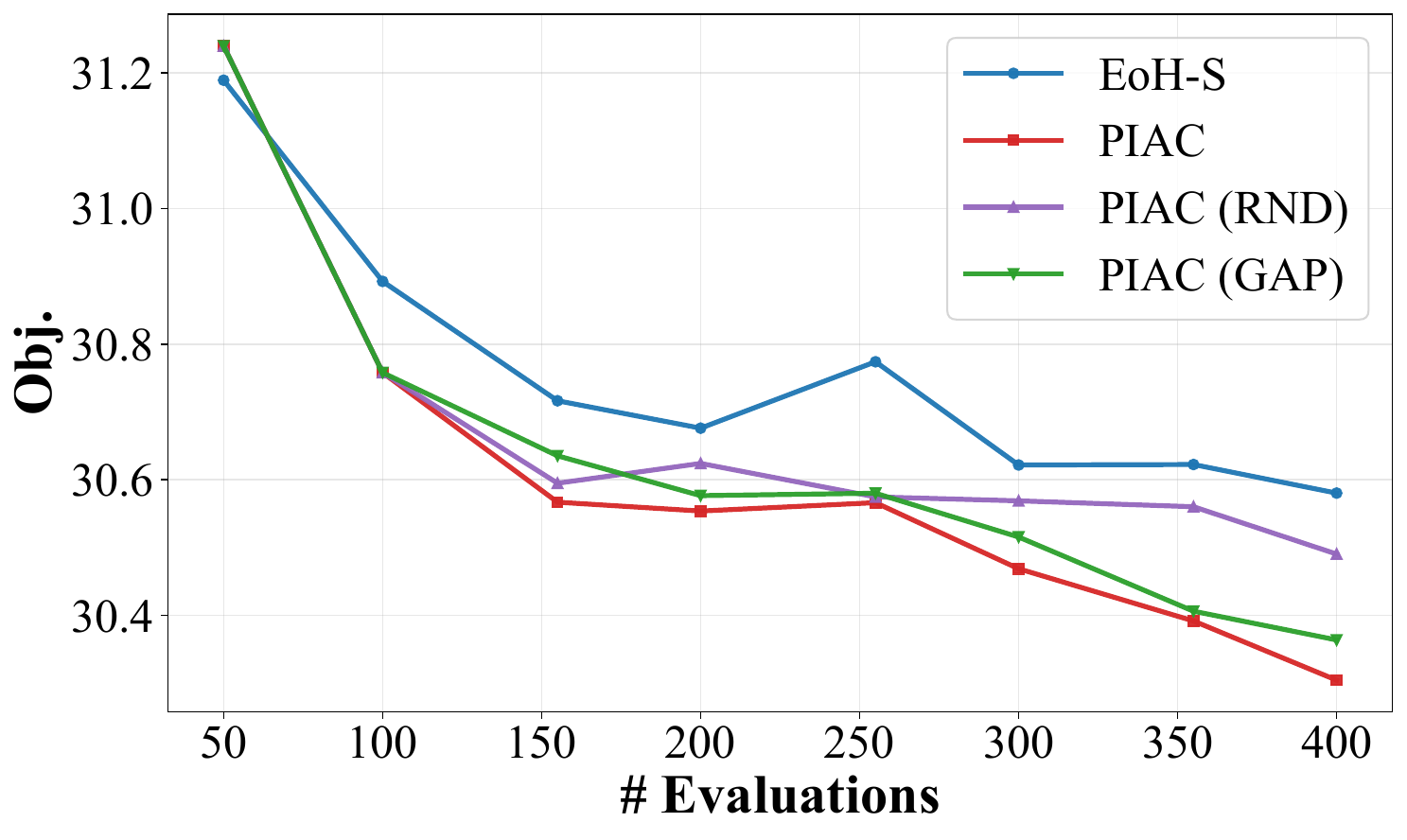}\\[-1mm]
        {\small (d) CVRP-Cluster}
    \end{minipage}

    \caption{Performance curves under different numbers of algorithm evaluations. Lower objective values indicate better performance.}
    \label{fig:evaluation-curves}
\end{figure*}

\subsection{Performance Dynamics of the Algorithm Portfolio}
\label{sec:rq3}

To answer \emph{RQ4} regarding whether the introduction of newly generated valuable instances can continuously improve performance, we track the performance dynamics of the algorithm portfolios throughout the evolution process for the TSP constructive and CVRP constructive heuristics. In our experimental setup, the portfolio is initially trained on the initial dataset of 8 instances. Subsequently, PIAC dynamically expands the training dataset by introducing 8 newly synthesized problem instances every 100 algorithm evaluations ($FE$). Figure~\ref{fig:evaluation-curves} illustrates the performance of the algorithm portfolios across different test data distributions, including TSP-Rue, TSP-Cluster, CVRP-Rue, and CVRP-Cluster.

On the TSP test datasets (TSP-Rue and TSP-Cluster), a clear divergence in search dynamics emerges. During the initial 100 algorithm evaluations, EoH-S and PIAC achieve highly comparable performance, as both methods rapidly reduce the objective values. However, the improvement of EoH-S slows down significantly thereafter, reaching a performance plateau after 200 evaluations with limited further gains. In contrast, PIAC sustains a continuous downward trend throughout the entire evaluation process. By dynamically introducing valuable training data, PIAC effectively avoids early convergence and consistently refines the algorithm portfolio.
Transitioning to the CVRP test datasets, although both EoH-S and PIAC achieve substantial performance gains on CVRP-Rue, PIAC maintains a higher rate of improvement. Furthermore, a distinct divergence occurs in the later stages across different distributions. While EoH-S continues to achieve considerable gains on CVRP-Rue after 200 evaluations, its improvement on the CVRP-Cluster distribution becomes severely limited. A similar stagnation is observed for PIAC (RND) on CVRP-Cluster. This limitation arises because PIAC (RND) introduces instances from a fixed random distribution, failing to adequately cover the structural properties of the Cluster distribution. In contrast, by leveraging LLMs to construct diverse instance mutators, both PIAC (GAP) and the full PIAC ensure that the generated training data covers a significantly broader problem space, enabling them to continuously enhance performance on the Cluster distribution.
In summary, the evaluation dynamics explicitly answer \emph{RQ4} and confirm that continuously introducing newly generated valuable instances successfully drives the sustained performance improvement of the algorithm portfolio.

\subsection{Foundation Model Analysis}

To analyze PIAC's performance across different LLMs, we evaluate PIAC using DeepSeek-V3.2~\cite{deepseekv3}, DeepSeek-V4~\cite{deepseekv4}, Kimi-K2.6~\cite{kimi2}, and GPT-4.1 mini~\cite{gpt4}. As shown in Table~\ref{tab:llm_compare}, the quality of the constructed algorithm portfolios consistently improves as the capability of the underlying model increases. DeepSeek-V4 achieves the best overall performance, reducing the optimality gaps to 10.08 and 10.31 on the Rue and Cluster instances, respectively.
\begin{table}[htbp]
\centering
\caption{Performance comparison of different LLMs on Rue and Cluster instances using PIAC.}
\label{tab:llm_compare}
\begin{tabular}{lcc}
\toprule
\textbf{LLM} & \textbf{Rue} & \textbf{Cluster} \\
\midrule
DeepSeek-V3.2~\cite{deepseekv3} & 11.54\% & 12.23\% \\
DeepSeek-V4~\cite{deepseekv4} & \textbf{10.08\%} & \textbf{10.31\%} \\
Kimi-K2.6~\cite{kimi2} & 17.72\% & 17.68\% \\
GPT-4.1 mini~\cite{gpt4} & 13.63\% & 16.09\% \\
\bottomrule
\end{tabular}
\end{table}

\begin{table}[t]
\centering
\caption{
Ablation study on complementary operation.
Lower values are better.
Values in parentheses indicate the performance degradation compared with PIAC.
}
\label{tab:ablation_selection}
\begin{tabular}{lcc}
\toprule
\textbf{Method}
& \textbf{Rue}
& \textbf{Cluster} \\
\midrule

EoH-S
& 18.71\%
& 19.07\% \\

\midrule

\makecell[l]{PIAC\\ w/ random}
& \makecell{17.52\%\\(+1.63\%)}
& \makecell{18.52\%\\(+2.25\%)} \\

\midrule

\makecell[l]{PIAC\\ w/ distance}
& \makecell{16.47\%\\(+0.58\%)}
& \makecell{17.04\%\\(+0.77\%)} \\

\midrule

PIAC
& \textbf{15.89\%}
& \textbf{16.27\%} \\

\bottomrule
\end{tabular}

\end{table}
\subsection{Ablation Study: Complementary Operation}
\label{sec:ablation_parent_selection}

We conduct an ablation study to evaluate the effectiveness of the proposed Complementary Operation in guiding the evolutionary PAP algorithm. The experiments are performed on a fixed training dataset with 100 algorithm evaluations per portfolio. We compare three strategies: (i) \textit{random crossover operation}, where parent algorithms are randomly paired from the current algorithm pool; (ii) \textit{distance-based crossover operation}, where parent pairing is guided by score-list distance; and (iii) the PIAC method with the proposed complementary crossover operation, which integrates two-stage parent selection and complementary LLM prompting.

As shown in Table~\ref{tab:ablation_selection}, the PIAC variant using random crossover operation exhibits a clear performance degradation compared with the original PIAC, with Rue increasing from 15.89\% to 17.52\% and Cluster from 16.27\% to 18.52\%. The distance-based crossover operation improves over the random variant but still underperforms the original PIAC, indicating that score-list distance only captures partial diversity and fails to fully exploit complementary strengths. The original PIAC, leveraging the proposed complementary crossover operation with two-stage parent pairing and complementary LLM prompting, consistently achieves the best results. This demonstrates that explicitly guiding LLM-based crossover with complementary parent algorithms produces more diverse and synergistic heuristic rules, leading to superior algorithm portfolios.

\section{Conclusion}
\label{sec:conclusion}

This work introduced PIAC, a potential-aware co-evolutionary framework for automated algorithm portfolio construction. PIAC advances both problem-instance evaluation and generation through the potential gain metric and diverse LLM-evolved instance mutators, respectively. Specifically, by perturbing generated algorithms, potential gain quantifies achievable performance gains to identify instances with high quality, thereby bolstering portfolio generalization without relying on high-quality reference solutions. Simultaneously, the framework leverages LLMs to synthesize diverse instance mutators, expanding coverage across the problem space beyond hand-crafted operators. Together, these components guide the co-evolution of training instances and complementary portfolio members.
Extensive experiments on the TSP and CVRP across six data distributions demonstrate that PIAC consistently outperforms state-of-the-art LLM-based portfolio construction methods.
Ultimately, PIAC successfully extends the co-evolutionary paradigm into the domain of LLM-driven automated algorithm portfolio construction. While the current perturbation strategy assumes heuristic algorithms produce matrix-structured outputs, future research will design more universal perturbation schemes to accommodate arbitrary decision representations.

% \section*{Acknowledgments}
% This should be a simple paragraph before the References to thank those individuals and institutions who have supported your work on this article.

% \input{sections/appendix}

\bibliographystyle{IEEEtran}
\bibliography{main}

\IfFileExists{supplement.pdf}{
    \clearpage
    \includepdf[pages=-]{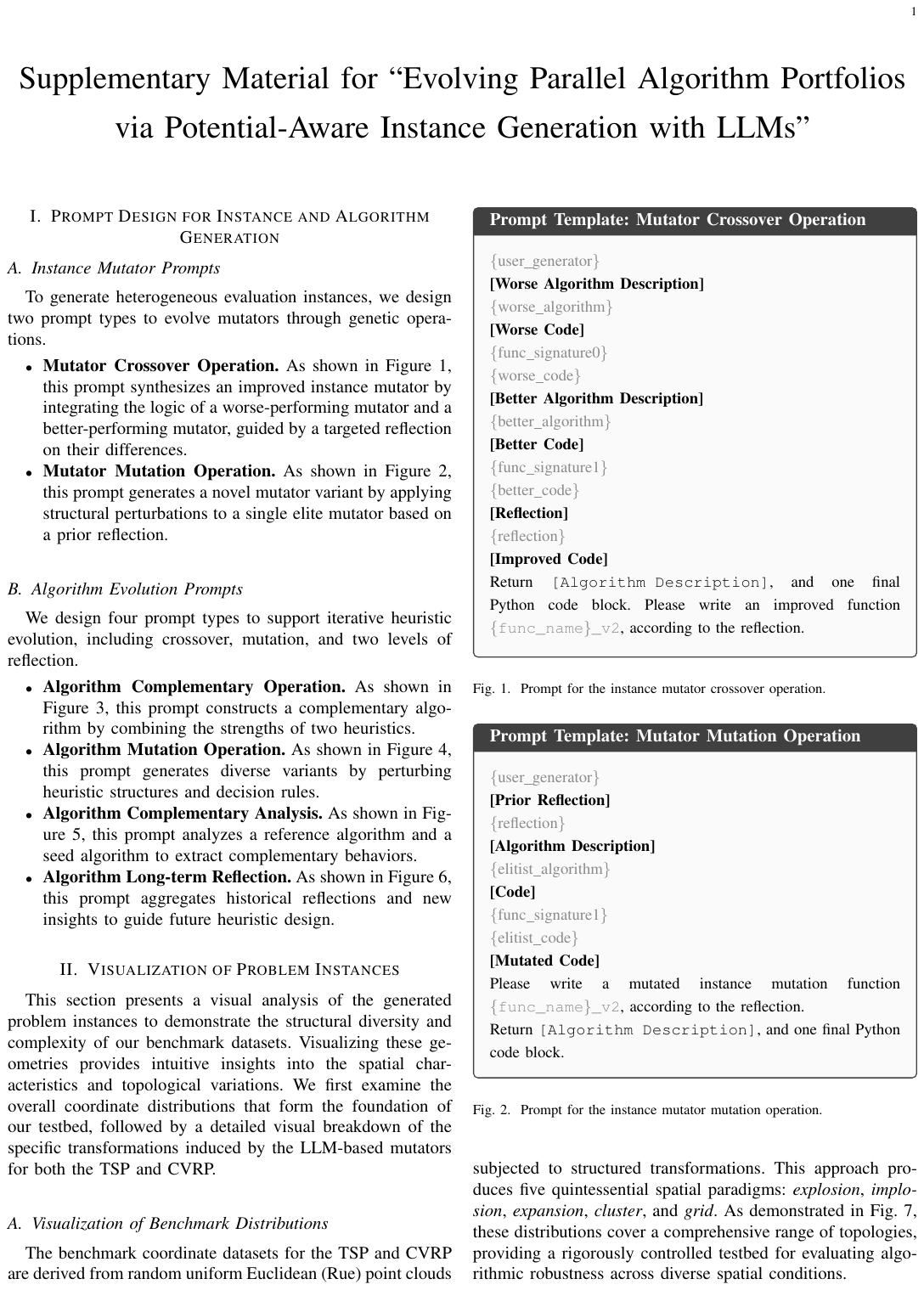}
}{
    \typeout{WARNING: supplement.pdf not found. Compile supplement.tex first to append it to main.pdf.}
}

\end{document}